\documentclass{svjour3}
\journalname{Machine learning}

\usepackage[numbers]{natbib}

\usepackage[sfdefault,lf]{carlito}
\usepackage[T1]{fontenc}

\usepackage{mathtools}
\usepackage{inconsolata}
\usepackage{booktabs}
\usepackage{listings}

\usepackage[misc]{ifsym}

\usepackage[utf8]{inputenc}

\newcommand{\metaho}{Metagol$_{ho}$}
\newcommand{\metabias}{Metagol$_{DF}$}
\newcommand{\forgetgol}{Forgetgol}
\newcommand{\poppi}{POPPI}

\newcommand{\dilp}{$\partial$ILP}
\newcommand{\tw}[1]{\texttt{#1}}
\usepackage{xcolor}

\newenvironment{code}
{
\ttfamily
\begin{center}
\begin{tabular}{|l|}
\hline
}
{
\\\hline
\end{tabular}
\end{center}
\par
}

\title{Inductive logic programming at 30}
\author{Andrew Cropper \and Sebastijan Dumančić \and Richard Evans \and Stephen H. Muggleton}
\institute{A. Cropper \Letter \at
              University of Oxford\\
              \email{andrew.cropper@cs.ox.ac.uk}
            \and
            S. Dumančić\at
              KU Leuven\\
              \email{sebastijan.dumancic@cs.kuleuven.be}
            \and
                R. Evans \at
                Imperial College London\\
              \email{richardevans@google.com}
            \and
                S. H. Muggleton \at
                Imperial College London\\
              \email{s.muggleton@imperial.ac.uk}
}

\begin{document}

\maketitle

\begin{abstract}
Inductive logic programming (ILP) is a form of logic-based machine learning.
The goal is to induce a hypothesis (a logic program) that generalises given training examples.
% and background knowledge.
As ILP turns 30, we review the last decade of research.
We focus on (i) new meta-level search methods, (ii) techniques for learning recursive programs, (iii) new approaches for predicate invention, and (iv) the use of different technologies.
We conclude by discussing current limitations of ILP and directions for future research.
\end{abstract}

\section{Introduction}
Inductive logic programming (ILP) \cite{mugg:ilp,mugg:ilp94} is a form of machine learning (ML).
As with other forms of ML, the goal is to induce a hypothesis that generalises training examples.
However, whereas most forms of ML use vectors/tensors to represent data, ILP uses logic programs (sets of logical rules).
    % Table-based learning is \emph{attribute-value} learning.
    % See \citet{luc:book} for an overview of the hierarchy of representations.
Moreover, whereas most forms of ML learn functions, ILP learns relations.

To illustrate ILP\footnote{
We do not introduce ILP in detail and refer the reader to the introductory paper of \citet{ilpintro} or the textbooks of \citet{ilp:book} and \citet{luc:book}.} suppose you want to learn a string transformation program from the following examples:

\begin{center}
\begin{tabular}{@{}c|c@{}}
\toprule
\textbf{Input} & \textbf{Output} \\
\midrule
inductive & e \\
logic & c \\
programming & g \\
\bottomrule
\end{tabular}
\end{center}

\noindent
Most forms of ML would use a table to represent these examples.
Each row would be an example.
Each column would be a feature, such as a one-hot-encoding representation of the string.
By contrast, in ILP, we would represent these examples as logical atoms, such as \tw{f([i,n,d,u,c,t,i,v,e], e)}, where \tw{f} is the target predicate that we want to learn (the relation to generalise).
We would also provide auxiliary information as background knowledge (BK), also represented as a logic program.
For instance, we could provide BK that contains logical definitions for string operations, such as
\tw{empty(A)}, which holds when the list \tw{A} is empty;
\tw{head(A,B)}, which holds when \tw{B} is the head of the list \tw{A};
and \tw{tail(A,B)}, which holds when \tw{B} is the tail of the list \tw{A}.
Given the aforementioned examples and BK, an ILP system could induce the hypothesis (a logic program):

\begin{code}
f(A,B):- tail(A,C),empty(C),head(A,B).\\
f(A,B):- tail(A,C),f(C,B).
\end{code}

\noindent
Each line of the program is a rule.
The first rule says that the relation \tw{f(A,B)} holds when the three literals \tw{tail(A,C)}, \tw{empty(C)}, and \tw{head(A,B)} hold.
In other words, the first rule says that \tw{B} is the last element of \tw{A} when the tail of \tw{A} is empty and \tw{B} is the head of \tw{A}.
The second rule is recursive and says that the relation \tw{f(A,B)} holds when the two literals \tw{tail(A,C)} and \tw{f(C,B)} hold.
In other words, the second rule says that \tw{f(A,B)} holds when the same relation holds for the tail of \tw{A}.

% \ac{@everyone, what do you think about adding a second traditional non-recursive ILP example, such as IGGP?}

\subsection{Why ILP?}

Compared to most ML approaches, ILP has several attractive features \cite{ilp30,ilpintro}:

\paragraph{Data efficiency.}
Many forms of ML are notorious for their inability to generalise from small numbers of training examples, notably deep learning \cite{marcus:2018,chollet:2019}.
As \citet{dilp} point out, if we train a neural system to add numbers with 10 digits, it might generalise to numbers with 20 digits, but when tested on numbers with 100 digits, the predictive accuracy drastically decreases  \cite{nandopoo,DBLP:journals/corr/KaiserS15}.
By contrast, ILP can induce hypotheses from small numbers of examples, often from a single example \cite{metabias,mugg:vision}.
% This data-efficiency is important because we often only have small amounts of training data.

\paragraph{Background knowledge.}
% In contrast to most forms of ML, which learn using finite tables of examples and features,
ILP learns using BK represented as a logic program.
Using logic programs to represent data allows ILP to learn with complex relational information, such as constraints about causal networks \cite{inoue:mla}, the axioms of the event calculus when learning to recognise events \cite{iled,oled}, and using a theory of light to understand images \cite{mugg:vision}.
Moreover, because hypotheses are symbolic, hypotheses can be added to BK, and thus ILP systems naturally support lifelong and transfer learning \cite{metabias,playgol,forgetgol}.

\paragraph{Expressivity.}
% ILP induces logic programs, which has many benefits.
% Because they are closely related to relational databases, logic programs naturally support relational data such as graphs.
Because of the expressivity of logic programs, ILP can learn complex relational theories, such as cellular automata \cite{inoue:lfit,apperception}, event calculus theories \cite{iled,oled}, Petri nets \cite{DBLP:journals/ml/BainS18}, answer set programs (ASP) \cite{ilasp}, and general algorithms \cite{popper}.
Because of the symbolic nature of logic programs, ILP can reason about hypotheses, which allows it to learn \emph{optimal} programs, such as minimal time-complexity programs \cite{metaopt} and secure access control policies \cite{law:fastlas}.

\paragraph{Expainability.}
Because of logic's similarity to natural language, logic programs can be easily read by humans, which is crucial for explainable AI.
For instance, \citet{mugg:compmlj} provide the first demonstration of \emph{ultra-strong ML} \cite{usml}, where a learned hypothesis is expected to not only be accurate but to also demonstrably improve the performance of a human when provided with the learned hypothesis.

\subsection{Recent advances}
Some of the aforementioned advantages come from developments in the last decade of ILP research, which we survey in this paper\footnote{
    This paper extends the paper of \citet{ilp30}.
}.
To aid the reader, we coarsely compare old and new ILP systems, where new represents systems from the past decade.
We use FOIL \cite{foil}, Progol \cite{progol}, Aleph \cite{aleph}, TILDE \cite{tilde}, and HYPER \cite{hyper} as representative old systems and ILASP \cite{ilasp}, Metagol \cite{metagol}, \dilp{} \cite{dilp}, and Popper \cite{popper} as representative new systems.
This comparison, shown in Table \ref{tab:diffs}, is, of course, vastly oversimplified, and there are many exceptions.
% \ac{make it clearer to R1 that there are exceptions}
In the rest of this paper, we survey these developments (each row in the table) in turn.
After discussing these new ideas, we discuss recent application areas (Section \ref{sec:apps}) before concluding by proposing directions for future research.

\begin{table}[ht]
\centering
\begin{tabular}{@{}c|c|c@{}}
\toprule
& \textbf{Old ILP} & \textbf{New ILP} \\
\midrule
\textbf{Search method} & Top-down and Bottom-up & Meta-level\\
\midrule
\textbf{Recursion} & Limited & Yes\\
\midrule
\textbf{Predicate invention} & No & Limited\\
\midrule
\textbf{Hypotheses} & First-order & ASP, Higher-order, Probabilistic\\
\midrule
\textbf{Optimality} & No & Yes\\
\midrule
\textbf{Technology} & Prolog & Prolog, ASP, NNs \\
\bottomrule
\end{tabular}
\caption{
    A simplified comparison of old and new ILP systems.
}
\label{tab:diffs}
\end{table}

% \subsection{ILP problem setting}
% We now briefly introduce the most popular ILP learning setting\footnote{
% As \citet{luc:settings} states, there are three main ILP learning settings: learning from \emph{entailment}, \emph{interpretations}, and \emph{satisfiability}. Most ILP approaches learn from entailment or interpretations.  Some recent work focuses on \emph{learning from transitions} \cite{inoue:lfit,apperception,TRMLJ2020}.We refer the reader to those works for an overview of that new learning setting.}: learning from entailment \cite{luc:book}

% \begin{definition}[\textbf{Learning from entailment}]
% \label{def:lfe}
% Given a tuple $(B,E^+,E^-,\mathcal{L})$ where:
% \begin{itemize}
%     \setlength\itemsep{1pt}
%     \setlength\parskip{1pt}
% \item $B$ denotes background knowledge
% \item $E^{+}$ denotes positive examples of the concept
% \item $E^{-}$ denotes negative examples of the concept
% \end{itemize}
% The goal LFE is to return a hypothesis $H \in \mathcal{H}_{\mathcal{L}}$ such that:
% \begin{itemize}
%     \setlength\itemsep{1pt}
%     \setlength\parskip{1pt}
%     \item $\forall e \in E^+, \; H \cup B \models e$ (i.e.~$H$ is \emph{complete})
%     \item $\forall e \in E^-, \; H \cup B \not\models e$ (i.e.~$H$ is \emph{consistent})
% \end{itemize}
% \noindent
% \end{definition}
\section{Search methods}
\label{sec:method}

The fundamental ILP problem is to efficiently search a large hypothesis space.
Most older ILP approaches search in either a \emph{top-down} or \emph{bottom-up} fashion.
These methods rely on notions of generality (typically theta-subsumption \cite{plotkin:thesis}), where one program is more \emph{general} or more \emph{specific} than another.
A third search approach has recently emerged called \emph{meta-level} ILP \cite{inoue:mla,mugg:metagold,inoue:flap,law:alp,popper}.
We discuss these approaches in turn.

\subsection{Top-down and bottom-up}

Top-down approaches \cite{foil,tilde,hyper} start with a general hypothesis and then specialise it.
HYPER, for instance, searches a tree in which the nodes correspond to hypotheses and each child of a hypothesis in the tree is more specific than or equal to its predecessor in terms of theta-subsumption.
% , i.e.~a hypothesis can only entail a subset of the examples entailed by its parent.
An advantage of top-down approaches is that they can often learn recursive programs (although not all do).
A disadvantage is that they can be prohibitively inefficient because they can generate many hypotheses that do not cover the examples.

Bottom-up approaches, by contrast, start with the examples and generalise them \cite{duce,cigol,golem,inoue:lfit}.
For instance, Golem \cite{golem} generalises pairs of examples based on relative least-general generalisation \cite{ilp:book}.
Bottom-up approaches can be seen as being \emph{data-} or \emph{example-driven}.
An advantage of these approaches is that they are typically fast.
As \citet{hyper} points out, disadvantages include (i) they typically use unnecessarily long hypotheses with many clauses, (ii) it is difficult for them to learn recursive hypotheses and multiple predicates simultaneously, and (iii) they do not easily support predicate invention.

Progol \cite{progol}, which inspired many other ILP approaches \cite{aleph,xhail,atom,inspire}, combines both top-down and bottom-up approaches.
Starting with an empty program, Progol picks an uncovered positive example to generalise.
To generalise an example, Progol uses mode declarations to build the \emph{bottom clause} \cite{progol}, the logically most-specific clause that explains the example.
The bottom clause bounds the search from below (the bottom clause) and above (the empty set).
Progol then uses an A* algorithm to generalise the bottom clause in a top-down (general-to-specific) manner and uses the other examples to guide the search.
% Although powerful, Progol struggles to learn recursive and optimal programs and does not support predicate invention.

\subsubsection{Meta-level}

Top-down and bottom-up approaches refine and revise a single hypothesis.
A third approach has recently emerged called \emph{meta-level} ILP \cite{inoue:mla,mugg:metagold,inoue:flap,law:alp,popper,patmug:topprog}.
There is no standard definition for \emph{meta-level ILP}.
Most approaches encode the ILP problem as a meta-level logic program, i.e.~a program that reasons about programs.
Meta-level approaches then often delegate the search for a hypothesis to an off-the-shelf solver \cite{aspal,metagol,ilasp,hexmil,inspire,apperception,popper} after which the meta-level solution is translated back to a standard solution for the ILP task.
In other words, instead of writing a procedure to search in a top-down or bottom-up manner, most meta-level approaches formulate the learning problem as a declarative search problem.
For instance, ASPAL \cite{aspal} translates an ILP task into a meta-level ASP program that describes every example and every possible rule in the hypothesis space.
ASPAL then delegates the search to an ASP system to find a subset of the rules that covers all the positive but none of the negative examples.

The main advantage of meta-level approaches is that they can more easily learn recursive programs and optimal programs \cite{aspal,ilasp,metagol,hexmil,apperception,popper}, which we discuss in Sections \ref{sec:recursion} and \ref{sec:optimal} respectively.
% Meta-level approaches can often learn optimal and recursive programs.
Moreover, whereas classical ILP systems were almost entirely based on Prolog, meta-level approaches use diverse techniques and technologies, such as ASP solvers \cite{aspal,ilasp,hexmil,popper,apperception}, which we expand on in Section \ref{sec:tech}.
 % and the Apperception Engine \cite{apperception}  translate an ILP problem into an ASP problem and use powerful ASP solvers to find a model of the problem -- note that these systems all employ very different algorithms.
The development of meta-level ILP approaches has, therefore, diversified ILP from the standard clause refinement approach of earlier ILP systems.

Most meta-level approaches encode the ILP learning task as a single static meta-level program \cite{aspal,ilasp,hexmil,apperception}.
 % ASPAL and HEXMIL translate an ILP task into a single meta-level ASP program that they pass to an ASP solver.
% that it then hands off to an ASP solver.
A major issue with this approach is that the meta-level program can be very large so these approaches can struggle to scale to problems with non-trivial domains and programs with large clauses.
 % \cite{metaho,popper}.
Two related approaches try to overcome this limitation by continually revising the meta-level program.

ILASP3 \cite{ilasp3} employs a counter-example-driven select-and-constrain loop.
ILASP3 first pre-computes every clause in the hypothesis space defined by a set of given mode declarations \cite{progol}.
ILASP3 then starts its select-and-constrain loop.
With each iteration, ILASP3 uses an ASP solver to find the best hypothesis (a subset of the rules) it can.
If the hypothesis does not cover one of the examples, ILASP3 finds a reason why and then generates constraints (boolean formulas over the rules) which it adds to the meta-level program to guide subsequent search.
% \footnote{
% This statement covers the noiseless ILASP3 setting.
% Things are slightly more complicated in noisy tasks where examples are given penalties and ILASP3 may return a hypothesis that does not cover all examples, but is optimal with respect to the penalties.
% }.
Another way of viewing ILASP3 is that it uses a counter-example-guided approach and translates an uncovered example $e$ into a constraint that is satisfied if and only if $e$ is covered.
% ILASP3's constraint loop requires at most $|E|$ iterations, where $|E|$ is the number of ILASP3 examples, which are partial interpretations.

Popper \cite{popper} adopts a similar approach but differs in that it (i) does not precompute every possible rule in the hypothesis space, and (ii) translates a \emph{hypothesis}, rather than an uncovered example, into a set of constraints.
% footnote{
% ILASP3's loop requires at most $|E|$ iterations, where $|E|$ is the number of examples.
% % , which are partial interpretations.
% % Given only a single example, ILASP3 would only loop once.
% Popper's loop requires at most $|\mathcal{H}|$ iterations, where $|\mathcal{H}|$ is the size of the hypothesis space.
% % regardless of the number of examples.
% }.
Popper works in three repeating stages: \emph{generate}, \emph{test}, and \emph{constrain}.
Popper first constructs a meta-level logic program where its models correspond to hypotheses.
In the generate stage, Popper asks an ASP solver to find a model (a hypothesis).
% program that satisfies a set of \emph{hypothesis constraints}.
In the test stage, Popper tests the hypothesis against the examples.
A hypothesis \emph{fails} when it is incomplete (does not entail all the positive examples) or inconsistent (entails a negative example).
If a hypothesis fails, Popper learns constraints from the failure, which it then uses to restrict subsequent generate stages.
For instance, if a hypothesis is inconsistent, then Popper generates a generalisation constraint to prune all generalisations of the hypothesis and adds the constraint to the meta-level program, which eliminates models and thus prunes the hypothesis space.
This process repeats until Popper finds a complete and consistent program.

% Although meta-level approaches have many advantages over the classical top-down and bottom-up approaches, some unresolved issues remain.
% Since most ASP solvers only work on ground programs \cite{clingo}, pure ASP-based approaches are inherently restricted to tasks that have a small and finite grounding.
% Although preliminary work attempts to tackle this issue \cite{popper}, work is still needed for these approaches to scale to very large problems.
% Many approaches also precompute every possible rule in a hypothesis \cite{aspal,ilasp}, so struggle to learn programs with large rules, although preliminary work tries to address this issue \cite{brute}.

For more information about meta-level learning, we suggest the work of \citet{inoue:flap} and \citet{law:alp}.

\section{Recursion}
\label{sec:recursion}

Learning recursive programs has long been considered a difficult problem for ILP \cite{ilp20,ilpintro}.
The power of recursion is that an infinite number of computations can be described by a finite recursive program \cite{DBLP:books/daglib/0067086}.
To illustrate the importance of recursion, reconsider the string transformation problem from the introduction.
Without recursion, an ILP system would need to learn a separate clause to find the last element for each list of length $n$, such as this program for when $n=3$:

\begin{code}
f(A,B):- tail(A,C),empty(C),head(A,B).\\
f(A,B):- tail(A,C),tail(C,D),empty(D),head(C,B).\\
f(A,B):- tail(A,C),tail(C,D),tail(D,E),empty(E),head(D,B).
\end{code}

\noindent
This program does not generalise to lists of arbitrary lengths.
Moreover, most ILP systems would need examples of lists of each length to learn such a program.
By contrast, an ILP system that supports recursion can learn the compact program:

\begin{code}
f(A,B):- tail(A,C),empty(C),head(A,B).\\
f(A,B):- tail(A,C),f(C,B).
\end{code}

\noindent
Because of the symbolic representation and the recursive nature, this program generalises to lists of arbitrary length and which contain arbitrary elements (e.g. integers and characters).
In general, without recursion, it can be difficult for an ILP system to generalise from small numbers of examples \cite{datacurate}.

Older ILP systems struggle to learn recursive programs, especially from small numbers of training examples.
A common limitation with existing approaches is that they rely on \emph{bottom clause} construction \cite{progol}.
In this approach, for each example, an ILP system creates the most specific clause that entails the example and then tries to generalise the clause to entail other examples.
However, this sequential covering approach requires examples of both the base and inductive cases.
The classical ILP system FOIL \cite{foil} also struggles to learn recursive programs because it induces programs one clause at a time.
% \ac{mention the counter-example of FOIL - on p.5 you claim that old ILP systems struggle with recursion, as they rely on bottom clause construction [77]. This is just partially justified, as e.g. even the early FOIL top-down relation learning system was able to learn recursive definitions, using the notion of an irreflexive partial ordering}.

Interest in recursion has resurged with the introduction of meta-interpretive learning (MIL) \cite{mugg:metalearn,mugg:metagold,metaho} and the MIL system Metagol \cite{metagol}.
The key idea of MIL is to use \emph{metarules} \cite{reduce}, or program templates, to restrict the form of inducible programs, and thus the hypothesis space\footnote{
    The idea of using metarules to restrict the hypothesis space has been widely adopted by many approaches \cite{wang2014structure,awscp,DBLP:conf/nips/Rocktaschel017,dilp,DBLP:journals/ml/BainS18,hexmil}.
    However, despite their now widespread use, there is little work determining which metarules to use for a given learning task (\cite{reduce} is an exception), which future work must address.
}.
A metarule is a higher-order clause.
For instance, the \emph{chain} metarule is $P(A,B) \leftarrow Q(A,C), R(C,B)$, where the letters $P$, $Q$, and $R$ denote higher-order variables and $A$, $B$ and $C$ denote first-order variables.
The goal of a MIL system, such as Metagol, is to find substitutions for the higher-order variables.
For instance, the \emph{chain} metarule allows Metagol to induce programs such as \tw{f(A,B):- tail(A,C),head(C,B)}\footnote{Metagol can induce longer clauses though predicate invention, which is described in Section \ref{sec:pi}.}.
Metagol induces recursive programs using recursive metarules, such as the \emph{tailrec} metarule \emph{P(A,B) $\leftarrow$ Q(A,C), P(C,B)}.

Following MIL, many meta-level ILP systems can learn recursive programs \cite{ilasp,dilp,hexmil,popper}.
With recursion, ILP systems can now generalise from small numbers of examples, often a single example \cite{metabias}.
Moreover, the ability to learn recursive programs has opened up ILP to new application areas, including learning string transformations programs \cite{metabias}, answer set grammars \cite{law:asg}, and general algorithms \cite{popper}.

\section{Predicate invention}
\label{sec:pi}

A key characteristic of ILP is the use of BK, which contains facts and rules (extensional and intensional definitions) in the form of a logic program.
For instance, when learning string transformation programs, we may provide helper background relations, such as \tw{head/2} and \tw{tail/2}.
For other domains, we may supply more complex BK, such as a theory of light to understand images \cite{mugg:vision} or higher-order operations, such as \tw{map/3}, \tw{filter/3}, and \tw{fold/4}, to solve programming puzzles \cite{metaho}.

Choosing appropriate BK is crucial for good learning performance.
ILP has traditionally relied on hand-crafted BK, often designed by domain experts.
This approach is limited because obtaining suitable BK can be difficult and expensive.
Indeed, the over-reliance on hand-crafted BK is a common criticism of ILP \cite{dilp}.

Rather than expecting a user to provide all the necessary BK, the goal of \emph{predicate invention} (PI) \cite{cigol,stahl:pi} is for an ILP system to automatically invent new auxiliary predicate symbols.
This idea is similar to when humans create new functions when manually writing programs, to reduce code duplication or to improve readability.
Whilst PI has attracted interest since the beginnings of ILP \cite{cigol}, and has subsequently been repeatedly stated as a major challenge \cite{pedro:pi,ilp20,kramer:ijcai20}, most ILP systems do not support it, including classical systems, such as Progol \cite{progol}, TILDE \cite{tilde}, and Aleph \cite{aleph}, and modern systems, such as ATOM \cite{atom} and LFIT \cite{inoue:lfit}.

A key challenge faced by ILP systems is deciding when and how to invent a new symbol.
As \citet{kramer1995predicate} points out, PI is difficult because it is unclear how many arguments an invented predicate should have, how the arguments should be ordered, etc.
Several PI approaches try to address this challenge, which we discuss in turn.

\subsection{Placeholders}
A classical approach to PI is to predefine invented symbols through mode declarations, which \citet{DBLP:conf/ilp/LebanZB08} call \emph{placeholders}.
However, this placeholder approach is limited because it requires that a user manually specify the arity and argument types of a symbol \cite{ilasp}, which rather defeats the point, or requires generating all possible invented predicates \cite{dilp,apperception}, which is computationally expensive.

\subsection{Metarules}
\label{sec:metarules}

Interest in \emph{automatic} PI (where a user does not need to predefine an invented symbol) has resurged with the introduction of MIL.
MIL avoids the issues of older ILP systems by using metarules to define the hypothesis space and in turn reduce the complexity of inventing a new predicate symbol.
For instance, the \emph{chain} metarule ($P(A,B) \leftarrow Q(A,C), R(C,B)$) allows Metagol to induce programs such as \tw{f(A,B):- tail(A,C),tail(C,D)}, which would drop the first two elements from a list.
To induce longer clauses, such as to drop first three elements from a list, Metagol uses the same metarule but invents a predicate symbol to chain their application, such as to induce the program:

\begin{code}
f(A,B):- tail(A,C),inv1(C,B).\\
inv1(A,B):- tail(A,C),tail(C,B).
\end{code}

\noindent
To learn this program, Metagol invents the predicate symbol \tw{inv1} and induces a definition for it using the \emph{chain} metarule.
Metagol uses this new predicate symbol in the definition for the target predicate \tw{f}.

A side-effect of this metarule-driven approach is that problems are forced to be decomposed into reusable solutions.
For instance, to learn a program that drops the first four elements of a list, Metagol learns the following program, where the invented predicate symbol \tw{inv1} is used twice:

\begin{code}
f(A,B):- inv1(A,C),inv1(C,B).\\
inv1(A,B):- tail(A,C),tail(C,B).
\end{code}

\noindent
PI has been shown to help reduce the size of target programs, which in turn reduces sample complexity and improves predictive accuracy \cite{playgol}.
Several new ILP systems support PI using a metarule-guided approach \cite{dilp,hexmil,celine:bottom}.
% The principle of introducing new predicate symbols without explicit human-guidance is known as \emph{automatic predicate invention}.
% \subsection{Failure-driven PI}
% \ac{talk about Popper}

\subsection{Pre/post-processing}
Metarule-driven PI approaches perform PI during the learning task.
A recent trend is to perform PI as a pre/post-processing step to improve knowledge representation \cite{curled,alps,playgol,celine:bottom}.

% \ac{@SM, short paragraph on celine's work}

% Given a set of user-supplied tasks, Playgol \cite{playgol} employs a two-stage approach.
% First, Plagol \emph{plays} by generating random tasks to solve  and tries to solve them, adding any solutions to the BK, which can be seen as a form of \emph{self-supervised} learning.
% After playing, Playgol then \emph{builds} by trying to solve the user-supplied tasks by reusing solutions learned whilst playing.
% The goal of Playgol is similar to all the approaches discussed in this section: to automatically discover reusable general programs to improve learning performance, but does so with fewer labelled examples.

CUR$^2$LED \cite{curled} performs PI by clustering constants and relations in the provided BK, turning each identified cluster into a new BK predicate.
The key insight of CUR$^2$LED is not to use a single similarity measure, but rather a set of various similarities.
This choice is motivated by the fact that different similarities are useful for different tasks, but in the unsupervised setting the task itself is not known in advance.
CUR$^2$LED performs PI by producing different clusterings according to the features of the objects, community structure, and so on.

ALPs \cite{alps} perform PI using an auto-encoding principle: they learn an \emph{encoding} logic program that maps the provided data to a new, compressive latent representation (defined in terms of the invented predicates), and a \emph{decoding} logic program that can reconstruct the provided data from its latent representation.
This approach shows improved performance on supervised tasks, even though the PI step is task-agnostic.

Knorf~\cite{knorf} pushes the idea of ALPs even further.
Knorf compresses a program by removing redundancies in it.
If the learnt program contains invented predicates, Knorf revises them and introduces new ones that would lead to a smaller program.
The refactored program is smaller in size and contains less redundancy in clauses, both of which lead to improved performance.
The authors experimentally demonstrate that refactoring improves learning performance in lifelong learning and that Knorf substantially reduces the size of the BK program, reducing the number of literals in a program by 50\% or more.

\subsection{Lifelong Learning}
\label{sec:lifelong}

% An advantage of a symbolic representation is that learned knowledge can be added to the BK.
An approach to acquiring BK is to learn it in a lifelong learning setting.
The general idea is to reuse knowledge gained from solving one problem to help solve a different problem.

\metabias{} is an ILP system \cite{metabias} which given a set of tasks, uses Metagol to try to learn a solution for each task using at most one clause.
If Metagol finds a solution for a task, it adds the solution to the BK and removes the task from the set.
\metabias{} then asks Metagol to find solutions for the rest of the tasks but can now (i) use an additional clause, and (ii) reuse solutions from previously solved tasks.
This process repeats until \metabias{} solves all the tasks or reaches a maximum program size.
In this approach, \metabias{} automatically identifies easier problems, learn programs for them, and then reuses the solutions to help learn programs for more difficult problems.
The authors experimentally show that their multi-task approach performs substantially better than a single-task approach because learned programs are frequently reused and leads to a hierarchy of induced programs.
% BK composed of reusable programs, where each builds on simpler programs.

\metabias{} saves all learned programs (including invented predicates) to the BK, which can be problematic because too much irrelevant BK is detrimental to learning performance \cite{ashwin:badbk}.
To address this problem, \forgetgol{} \cite{forgetgol} introduces the idea of \emph{forgetting}.
In this approach, \forgetgol{} continually grows and shrinks its hypothesis space by adding and removing learned programs to and from its BK.
The authors show that forgetting can reduce both the size of the hypothesis space and the sample complexity of an ILP learner when learning from many tasks.
% , which shows potential for ILP to be useful in a lifelong or continual learning setting, which is considered crucial for AI \cite{lake:ai}.

\subsection{Limitations}
The aforementioned techniques have improved the ability of ILP to invent high-level concepts.
However, PI is still difficult and there are many challenges to overcome, notably that (i) many systems struggle to perform PI at all, and (ii) those that do support PI mostly need much user-guidance, such as metarules to restrict the space of invented symbols or that a user specifies the arity and argument types of invented symbols.
There are notable exceptions.
\citet{DBLP:journals/jiis/Ferilli16a} describe an PI approach based on the ideal of specialising a theory to account for negative examples, similar to early work in non-monotonic ILP \cite{Bain92non-monotoniclearning}.
\poppi{} \cite{poppi} is an ILP system that supports automatic predicate invention, i.e. does not require metarules nor requires a user to predefine invented symbols.

By developing better approaches for PI, we can make progress on existing challenging problems.
For instance, in \emph{inductive general game playing} (IGGP) \cite{iggp}, the task is to learn the symbolic rules of games from observations of gameplay, such as learning the rules of \emph{connect four}.
The target solutions, which come from the general game playing competition \cite{ggp}, often contain auxiliary predicates.
For instance, the rules for \emph{connect four} are defined in terms of definitions for lines which are themselves defined in terms of columns, rows, and diagonals.
Although these auxiliary predicates are not strictly necessary to learn the target solution, inventing such predicates significantly reduces the size of the solution, which in turn makes them easier to learn.
Although new methods for PI can invent high-level concepts, they are not yet sufficiently powerful enough to perform well on the IGGP dataset.
Making progress in this area would constitute a major advancement in ILP.
% and a major step towards human-level AI.

\section{Hypotheses}
\label{sec:expr}

ILP systems have traditionally induced definite and normal logic programs, typically represented as Prolog programs.
A recent development has been to use different hypothesis representations.

\subsection{Datalog}
Datalog is a syntactical subset of Prolog which disallows complex terms as arguments of predicates and imposes restrictions on the use of negation.
 % (and negation with recursion).
% These restrictions make Datalog attractive for ILP for two reasons.
Datalog is a truly declarative language, whereas in Prolog reordering clauses can change the program.
Moreover, Datalog query is guaranteed to terminate, though this guarantee is at the expense of not being a Turing-complete language, which Prolog is.
Several works \cite{awscp,dilp,hexmil} induce Datalog programs.
The general motivation for reducing the expressivity of the representation language from Prolog to Datalog is to allow the problem to be encoded as a satisfiability problem, particularly to leverage recent developments in SAT and SMT.
We discuss the advantages of this approach more in Section \ref{sec:sat}.

\subsection{Answer Set Programming}
\label{sec:hyp:asp}
ASP \cite{asp} is a logic programming paradigm based on the stable model semantics of normal logic programs that can be implemented using the latest advances in SAT solving technology.
\citet{law:aij} discuss some of the advantages of learning ASP programs, rather than Prolog programs, which we reiterate.
When learning Prolog programs, the procedural aspect of SLD-resolution must be taken into account.
For instance, when learning Prolog programs with negation, programs must be stratified; otherwise, they may loop under certain conditions.
By contrast, as ASP is a truly declarative language, no such consideration needs to be taken into account when learning ASP programs.
Compared to Datalog and Prolog, ASP supports additional language constructs, such as disjunction in the head of a clause, choice rules, and hard and weak constraints.
A key difference between ASP and Prolog is semantics.
A definite logic program has only one model (the least Herbrand model).
By contrast, an ASP program can have one, many, or even no stable models (answer sets).
Due to its non-monotonicity, ASP is particularly useful for expressing common-sense reasoning \cite{ilasp3}.

To illustrate the benefits of learning ASP programs, we reuse an example from \citet{law:alp}.
Given a sufficient examples of Hamiltonian graphs, ILASP \cite{ilasp} can learn a program to definite them:

\begin{code}
0 \{ in(V0, V1) \} 1 :- edge(V0, V1).\\
reach(V0) :- in(1, V0).\\
reach(V1) :- reach(V0), in(V0, V1).\\
:- not reach(V0), node(V0).\\
:- V1 != V2, in(V0, V2), in(V0, V1).
\end{code}

\noindent
This program illustrates useful language features of ASP.
The first rule is a \emph{choice} rule, which means that an atom \emph{can} be true.
In this example, the rule indicates that there can be an in edge from the vertex \emph{V1} to \emph{V0}.
The last two rules are \emph{hard constraints}, which essentially enforce integrity constraints.
The first hard constraint states that it is impossible to have a node that is not reachable.
The second hard constraint states that it is impossible to have a vertex with two in edges from distinct nodes.
For more information about ASP we recommend the book by \citet{asp}.

Approaches to learning ASP programs can mostly be divided into two categories: \emph{brave learners}, which aim to learn a program such that at least one answer set covers the examples, and \emph{cautious learners}, which aim to find a program which covers the examples in all answer sets.
ILASP is notable because it supports both brave and cautious learning, which are both needed to learn some ASP programs \cite{law:aij}.
Moreover, ILASP differs from most Prolog-based ILP systems because it learns ASP programs, including programs with normal rules, choice rules, and both hard and weak constraints, which classical ILP systems cannot.
% \ac{- p. 10 please explain the reader what are unstratified ASP programs}.
Learning ASP programs allows for ILP to be used for new problems, such as inducing answer set grammars \cite{law:asg}.

% Comparing approaches that learn Prolog programs with those that learn ASP programs is, however, very difficult because of the language differences.
% For instance, Prolog does not support features such as choice rules or constraints and ASP frameworks do not support lists.

\subsection{Higher-order programs}
\label{sec:ho}

Imagine learning a \emph{droplasts} program, which removes the last element of each sublist in a list, e.g. \emph{[alice,bob,carol]} $\mapsto$ \emph{[alic,bo,caro]}.
Given suitable input data, Metagol can learn this first-order recursive program:

\begin{code}
f(A,B):- empty(A),empty(B).\\
f(A,B):- head(A,C),tail(A,D),head(B,E),tail(B,F),f1(C,E),f(D,F).\\
f1(A,B):- reverse(A,C),tail(C,D),reverse(D,B).
\end{code}

\noindent
Although semantically correct, the program is verbose.
To learn smaller programs, \metaho{} \cite{metaho} extends Metagol to support learning higher-order programs, where predicate symbols can be used as terms.
For instance, for the same \emph{droplasts} problem, \metaho{} learns the higher-order program:

\begin{code}
f(A,B):- map(A,B,f1).\\
f1(A,B):- reverse(A,C),tail(C,D),reverse(D,B).
\end{code}

\noindent
To learn this program, \metaho{} invents the predicate symbol \tw{f1}, which is used twice in the program: as term in the \tw{map(A,B,f1)} literal and as a predicate symbol in the \tw{f1(A,B)} literal.
Compared to the first-order program, this higher-order program is smaller because it uses \tw{map/3} (predefined in the BK) to abstract away the manipulation of the list and to avoid the need to learn an explicitly recursive program (recursion is implicit in \tw{map/3}).
\metaho{} has been shown to reduce sample complexity and learning times and improve predictive accuracies \cite{metaho}.
% This example illustrates the value of higher-order abstractions and inventions, which allow ILP systems to learn more complex programs using fewer examples with less search \cite{popho}.

\subsection{Probabilistic logic programs}
\label{sec:pilp}

A major limitation of logical representations, such as Prolog and its derivatives, is the implicit assumption that the BK is perfect.
That is, most ILP systems assume that atoms are true or false, leaving no room for uncertainty.
This assumption is problematic if data is noisy, which is often the case.

\begin{figure}[t]
    \centering
    \includegraphics[width=.9\linewidth]{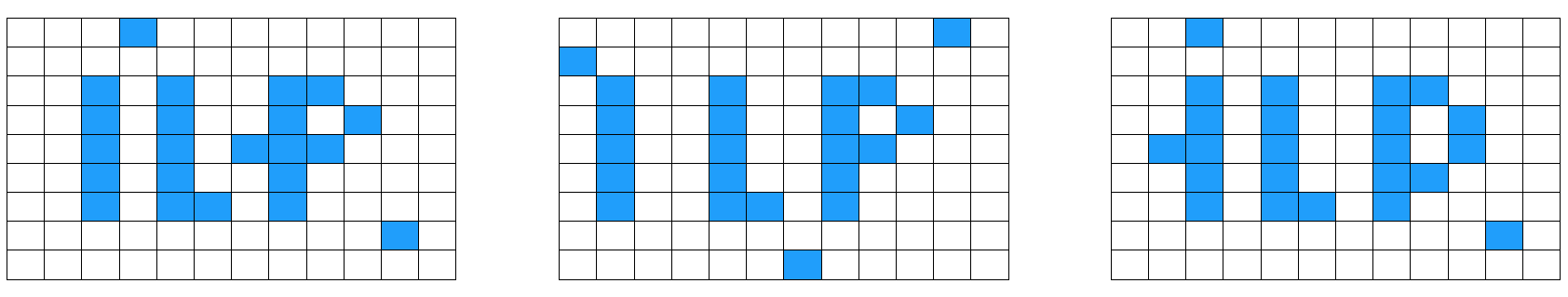}
    \caption{ILP systems struggle with structured examples that exhibit observational noise. All three examples clearly spell the word  "ILP", with some alterations: 3 noisy pixels, shifted and elongated letters. If we would be to learn a program that simply draws "ILP" in the middle of the picture, without noisy pixels and elongated letters, that would be a correct program. }
    \label{fig:noise}
\end{figure}

ILP systems have limited capabilities for dealing with noise. If a perfect program is not in the hypothesis space, the most common strategy is to find a program that covers as many positive and as few negative examples as possible. Though this approach helps us handle mislabelled examples, is not a good way for dealing with observational noise generally. The limitations of the approach become obvious when examples are structured (e.g., complex relations, images, or lists) rather than simple labels. Consider the example in Figure \ref{fig:noise}, in which we want to learn a program drawing simple images spelling "ILP".
Each of the examples displays a clear concept but with some alterations: every image contains three 'noisy' pixels: the second example has the first letter shifted to the left, while the arc in the letter "P" is elongated in the third example.
All of these alterations are noise and not something we want our program to explicitly represent - the ground truth program is the one that draws "ILP" in the middle of the figure, without any noise.
However, ILP systems based on entailment would consider a solution to be correct only if it models all of the noisy aspects.
An ILP system capable of handling such noise is Brute~\cite{brute}, which uses a distance to the target solution (e.g., pixel distance in case of images) as an optimisation criterion instead of entailment.

The most principle way of handling noise is to integrating probabilistic aspects into logical representations so that uncertainties in data can be directly modelled.
This integration is the focus of  statistical relational artificial intelligence (StarAI) \cite{pilp,DeRaedtKerstingEtAl16}.
In essence, StarAI hypothesis representations extend BK with probabilities or weights indicating the degree of confidence in the correctness of parts of BK.
StarAI is a big and prolific field; for that reason, we will not cover it in entirety but rather briefly introduce the main ideas that overcome limitations of logic programming-based ILP systems.

Generally, StarAI techniques are based on two ideas: \textit{distribution semantics} and \textit{maximum entropy}.
Distribution semantics approaches~\cite{Sato95astatistical}, including Problog~\cite{raedt:problog} and PRISM~\cite{Sato2001ParameterLO}, explicitly annotate uncertainties in BK.
To allow such annotation, they extend Prolog with two primitives for stochastic execution: probabilistic facts and annotated disjunctions.
Probabilistic facts are the most basic stochastic primitive and they take the form of logical facts labelled with a probability $p$.
Each probabilistic fact represents a Boolean random variable that is \tw{true} with probability $p$ and \tw{false} with probability $1-p$.
For instance, the following probabilistic fact states that there is 1\% chance of an earthquake in Naples.
\begin{code}
    0.01::earthquake(naples).
\end{code}
\noindent
An alternative interpretation of this statement is that 1\% of executions of the probabilistic program would observe an earthquake.
The second type of stochastic primitive is an annotated disjunction.
Whereas probabilistic facts introduce non-deterministic behaviour on the level of facts, annotated disjunctions introduce non-determinism on the level of clauses.
Annotated disjunctions allow for multiple literals in the head, where only one of the head literals can be \tw{true} at a time.
For instance, the following annotated disjunction states that a ball can be either green, red, or blue, but not a combination of colours:
\begin{code}
    $\frac{1}{3}$::colour(B,green); $\frac{1}{3}$::colour(B,red); $\frac{1}{3}$::colour(B,blue) :- ball(B).
\end{code}

\noindent
By contrast, maximum entropy approaches annotate uncertainties only at the level of a logical theory.
That is, they assume that the predicates in the BK are labelled as either true or false, but the label may be incorrect.
These approaches are not based on logic programming, but rather on first-order logic.
Consequently, the underlying semantics are different: rather than consider proofs, these approaches consider models or groundings of a theory.
This difference primarily changes what uncertainties represent.
For instance, Markov Logic Networks (MLN)~\cite{richardson:mln} represent programs as a set of weighted clauses.
The weights in MLN do not correspond to probabilities of a formula being true but, intuitively, to a log odds between a possible world (an interpretation) where the clause is true and a world where the clause is false.
For instance, a clause that is true in 80\% of the worlds would have a weight of 1.386 ($\log \frac{0.8}{0.2}$)

The techniques from learning such probabilistic programs are typically direct extensions of ILP techniques.
For instance, ProbFOIL~\cite{probfoil} extends FOIL \cite{foil} with probabilistic clauses.
Similarly, SLIPCOVER~\cite{DBLP:journals/tplp/BellodiR15} is a bottom-up approach, similar to Aleph~\cite{aleph} and Progol~\cite{progol}.
\citet{huynh:icml08}  use Aleph to find interesting clauses and then learn the corresponding weights.
\citet{kok:icml09} use relational pathfinding over BK to identify useful clauses.
That is, they interpret the BK as a hypergraph in which constants form vertices and atoms form hyper-edges and perform random walks.
Frequently occurring walks, or their subparts, are then turned into clauses.
Such random walks could be seen as an approximate way to construct bottom clauses.

It is worth noting that StarAI also considers an alternative learning problem - that of learning the probabilistic parameters of a given program.
We do not survey these approaches here as the problem is different in nature from the ILP problem: whereas ILP searches from a program solving the tasks, parameter learning methods assume that such program is given.

%All of these starAI framework require you to know the probability ahead of time
%This is not always reasonable - like images, or sounds
%
%DeepProblog extends Problog by allowing that the probabilities facts can be evaluated externally

\section{Optimality}
\label{sec:optimal}

There are often multiple (sometimes infinitely many) hypotheses that explain the data.
Deciding which hypothesis to choose has long been a difficult problem.
Many systems aim for maximum classification accuracy.
For instance, Aleph, by default, aims to maximum \emph{coverage} of each clause it adds to a hypothesis, where coverage is measured as $P - N$, where $P$ and $N$ are the number of positive and negative examples covered by the clause respectively.
Note that Aleph supports various evaluation metrics, such as \emph{compression}, measured as $P - N - L + 1$, where $P$ and $N$ are as before and $L$ is the number of literals in the clause.
However, older ILP systems are typically not guaranteed to induce optimal programs/theories.
A key reason for this limitation was that most search techniques learned a single clause at a time, leading to the construction of sub-programs that are sub-optimal in terms of program size and coverage.
For instance, Aleph offers no guarantee of optimality with respect to the program size and coverage.

Newer ILP systems try to address this limitation.
As with the ability to learn recursive programs, the main development is to take a global view of the induction task by using meta-level search techniques.
In other words, rather than induce a single clause at a time from a single example, the idea is to induce multiple clauses from multiple examples.
% For instance, ILASP precomputes every possible rule in the hypothesis space.
% The ILASP task is to find a minimal subset of clauses that covers as many positive and as few negative examples as possible.
For instance, ILASP uses ASP's optimisation abilities to provably learn the program with the fewest literals.
ILASP3 \cite{ilasp3} adopts a similar approach to account for noise.
% Likewise, other ILP systems are guaranteed to induce programs with the fewest clauses \cite{metagol,hexmil} or literals \cite{popper}.

% One claimed advantage of learning optimal programs is learning performance and that smaller program should provide better generalisations\footnote{
%     This advantage is not necessarily true. See the work of \citet{ilpintro} and \citet{pedro:occam} for discussions about learning textually minimal programs.
% }.
% When Law et al. \cite{law:noisy} compared ILASP (which is guaranteed to learn optimal programs) to Inspire \cite{inspire} (which is not guaranteed to learn optimal programs), ILASP achieved a higher F1 score (both systems were given identical hypothesis spaces and optimisation criteria).

% In addition to performance advantages, t
The ability to learn optimal programs opens up ILP to new problems.
For instance, learning efficient logic programs has long been considered a difficult problem in ILP \cite{mugg:ilp94,ilp20}, mainly because there is no declarative difference between an efficient program, such as mergesort, and an inefficient program, such as bubble sort.
To address this issue, Metaopt \cite{metaopt} extends Metagol to support learning efficient programs.
Metaopt maintains a cost during the hypothesis search and uses this cost to prune the hypothesis space.
To learn minimal time complexity logic programs, Metaopt minimises the number of resolution steps.
For instance, imagine trying to learn a \emph{find duplicate} program, which finds any duplicate element in a list e.g. \emph{[p,r,o,g,r,a,m]} $\mapsto$ \emph{r}, and \emph{[i,n,d,u,c,t,i,o,n]
} $\mapsto$ \emph{i}.
Given suitable input data, Metagol can induce the program:

\begin{code}
f(A,B):- head(A,B),tail(A,C),element(C,B).\\
f(A,B):- tail(A,C),f(C,B).
\end{code}

\noindent
This program goes through the elements of the list checking whether the same element exists in the rest of the list.
Given the same input, Metaopt induces the program:

\begin{code}
f(A,B):- mergesort(A,C),f1(C,B).\\
f1(A,B):- head(A,B),tail(A,C),head(C,B).\\
f1(A,B):- tail(A,C),f1(C,B).
\end{code}

\noindent
This program first sorts the input list and then goes through the list to check whether for duplicate adjacent elements.
Although larger, both in terms of clauses and literals, the program learned by Metaopt is more efficient $O(n \log n)$ than the program learned by Metagol $O(n^2)$.
Metaopt has been shown to learn efficient robot strategies, efficient time complexity logic programs, and even efficient string transformation programs.

FastLAS \cite{law:fastlas} is an ASP-based ILP system that takes as input a custom scoring function and computes an optimal solution with respect to the given scoring function when learning non-recursive programs without PI.
The authors show that this approach allows a user to optimise domain-specific performance metrics on real-world datasets, such as access control policies.

\section{Technologies}
\label{sec:tech}

Older ILP systems mostly use Prolog for reasoning.
Recent work considers using different technologies.

\subsection{Constraint satisfaction and satisfiability}
\label{sec:sat}
There have been tremendous recent advances in SAT \cite{satprogresss}.
To leverage these advances, much recent work in ILP uses related techniques, notably ASP \cite{aspal,mugg:metalearn,ilasp,iled,oled,inspire,hexmil,apperception,popper}.
The main motivations for using ASP are to leverage (i) the language benefits of ASP (Section \ref{sec:hyp:asp}), and (ii) the efficiency and optimisation techniques of modern ASP solvers, such as CLASP \cite{clasp}, which supports conflict propagation and learning.
With similar motivations, other approaches encode the ILP problem as SAT \cite{atom} or SMT \cite{awscp} problems.
These approaches have been shown able to reduce learning times compared to standard Prolog-based approaches.
However, some unresolved issues remain.
A key issue is that most approaches encode an ILP problem as a single (often very large) satisfiability problem.
These approaches therefore often struggle to scale to very large problems \cite{metaho}, although preliminary work attempts to tackle this issue \cite{popper}.

\subsection{Neural networks}
With the rise of deep learning, several approaches have explored using gradient-based methods to learn logic programs.
These approaches all replace discrete logical reasoning with a relaxed version that yields continuous values reflecting the confidence of the conclusion.

The various neural approaches can be characterised along four orthogonal dimensions.
The first dimension is whether the neural network implements forward or backward inference.
While some \cite{DBLP:conf/nips/Rocktaschel017} use backward (goal-directed) chaining with a neural implementation of unification, most approaches \cite{dilp,Cohen_NeuralLP,NLM} use forward chaining.
The second dimension is whether the network is designed for big data problems \cite{Cohen_NeuralLP,DBLP:conf/nips/Rocktaschel017} or for data-efficient learning from a handful of data items \cite{dilp}.
Few neural systems to date are capable of handling both big data and small data, with the notable exception of \cite{NLM}.
The third dimension is whether the neural system jointly learns embeddings (mapping symbolic constants to continuous vectors) along with the logical rules \cite{DBLP:conf/nips/Rocktaschel017}.
The advantage of jointly learning embeddings is that it enables fuzzy unification between constants that are similar but not identical.
The challenge for these approaches that jointly learn embeddings is how to generalize appropriately to constants that have not been seen at training time.
%Moreover, neural methods often require millions of examples \cite{NLM} to learn concepts that symbolic ILP is able to learn from just a few.
The fourth dimension is whether or not the neural system is designed to allow explicit human-readable logical rules to be extracted from the weights of the network.
While most neural ILP systems \cite{Cohen_NeuralLP,DBLP:conf/nips/Rocktaschel017,dilp} do produce explicit logic programs, some \cite{NLM} do not.
It is perhaps moot whether implicit systems that do not produce explicit programs count as ILP systems at all -- but note that even in the implicit neural systems, the weight sharing of the neural net is designed to achieve strong generalisation by performing the same computation on all tuples of objects.

Currently, most neural approaches to ILP require the use of metarules or templates to make the search space tractable and fail to support predicate
invention, recursion and abduction.
This severely limits the applicability of these approaches, as the user cannot always be expected to provide suitable and complete background knowledge and metarules for a new problem.
The only approach that avoids the use of metarules or templates is Neural Logic Machines \cite{NLM}, and the only one to fully integrate neural net learning with predicate invention, recursion, and abduction is Abductive Meta-Interpretive Learning \cite{muggzhou:neuralmil}.

%The research has primarily developed in three directions.
%The first concerns imitating logical reasoning with tensor calculus \cite{Cohen_NeuralLP,NLM}.
%These approaches represent predicates as binary tensors over the domain of constants and perform reasoning by chains of tensor products imitating as clause.
%The second concerns the relaxation of the subset selection problem \cite{dilp} in which the task of a neural network is to select a subset of clauses from a space of pre-defined clauses.
%The third, \emph{neural theorem provers} \cite{DBLP:conf/nips/Rocktaschel017} turn the learning problem towards learning to perform \emph{soft} unification, which unifies not only the matching symbols but also similar ones, from a fixed set of proofs.
%
%The major challenge of neural approaches is the inability to generalise beyond training data and data efficiency.
%The majority of these approaches \textit{embed} logical symbols, i.e. they replace symbols with vectors, and therefore  a learned model is unable to work with unseen constants.
%Moreover, neural methods often require millions of examples \cite{NLM} to learn concepts that symbolic ILP is able to learn from just a few.

\section{Applications}
\label{sec:apps}

We now survey recent application areas for ILP.

\paragraph{Scientific discovery.}
Perhaps the most prominent application of ILP is in scientific discovery.
ILP has, for instance, been used to identify and predict ligands (substructures responsible for medical activity)~\cite{Kaalia16} and infer missing pathways in protein signalling networks~\cite{inoue:mla}.
There has been much recent work on applying ILP in ecology \cite{bohan2011automated,DBLP:conf/ilp/Tamaddoni-Nezhad14,bohan2017next}.
For instance, \citet{bohan2011automated} use ILP to generate plausible and testable hypotheses for trophic relations (`who eats whom') from ecological data.

% ILP is especially suitable for such problems because biological structures, including molecules and protein interaction networks, can easily be expressed as relations: molecular bonds define relations between atoms and interactions define relations between proteins.
% Moreover, as mentioned in the introduction, ILP induces human-readable models.
% ILP can, therefore, make predictions based on the (sub)structured present in biological structures which domain experts can interpret.
% \paragraph{Ecology.}
% \ac{@SM, can you add something?}

\paragraph{Program analysis.}
Due to the expressivity of logic programs as a representation language, ILP systems have found successful applications in software design.
ILP systems have proven effective in learning SQL queries \cite{awscp,Sivaraman2019}, programming language semantics \cite{DBLP:conf/ilp/BarthaC19}, and code search~\cite{Sivaraman2019}.

\paragraph{Robotics.}
\label{sec:robotics}
% Similarly to the previous category,
Robotics applications often require incorporating domain knowledge or imposing certain requirements on the learnt programs.
For instance, The Robot Engineer~\cite{SammutSHW15} uses ILP to design tools for robots and even complete robots, which are tests in simulations and real-world environments.
Metagol$_o$~\cite{metagolo} learns robot strategies considering their resource efficiency and \citet{AntanasMR15} recognise graspable points on objects through relational representations of objects.

\paragraph{Vision.}
Background knowledge is also valuable in Computer Vision.
Recent work
\citet{mugg:vision} demonstrated that
{\em Logical Vision}, which employs MIL, can outperform state-of-the-art
statistical machine learning in particular image recognition tasks,
given general Newtonian physics background knowledge concerning
reflection of light.

\paragraph{Games.}
% \ac{@SM, can you add to this?}
Inducing game rules has a long history in ILP, where chess has often been the focus \cite{chess-revision}.
\citet{DBLP:conf/ilp/LegrasRV18} show that Aleph and TILDE can outperform an SVM learner in the game of Bridge.
\citet{ilasp} use ILASP to induce the rules for Sudoku and show that this more expressive formalism allows for game rules to be expressed more compactly.
\citet{iggp} introduce the ILP problem of \emph{inductive general game playing}: the problem of inducing game rules from observations, such as \emph{Checkers}, \emph{Sokoban}, and \emph{Connect Four}. \citet{mugghoq:newgengamestrat} show the MIL system MIGO consistently outperforms deep reinforcement learning for both Noughts-and-Crosses and Hexapawn.

\paragraph{Data curation and transformation.}
Another successful application of ILP is in data curation and transformation, which is again largely because ILP can learn executable programs.
The most prominent example of such tasks is string transformations, such as the example given in the introduction.
There is much interest in this topic, largely due to success in synthesising programs for end-user problems, such as string transformations in Microsoft Excel \cite{flashfill}.
String transformations have become a standard benchmark for recent ILP papers~\cite{metabias,metaho,playgol,brute}.
Other transformation tasks include extracting values from semi-structured data (e.g. XML files or medical records), extracting relations from ecological papers, and spreadsheet manipulation~\cite{datacurate}.

\paragraph{Learning from trajectories.}
Learning from interpretation transitions (LFIT) \cite{inoue:lfit} automatically constructs a model of the dynamics of a system from the observation of its state transitions.
Given time-series data of discrete gene expression, it can learn gene interactions, thus allowing to explain and predict states changes over time \cite{TRMLJ2020}.
% \footnote{Implementations available at: \url{https://github.com/Tony-sama/pylfit}}.
LFIT has been applied to learn biological models, like Boolean Networks, under several semantics: memory-less deterministic systems \cite{inoue:lfit,DBLP:conf/ilp/RibeiroI14}, and their multi-valued extensions \cite{TRICMLA15,DMTRICAPS16}.
% For instaContinuous gene expression are tackled in \cite{TRILP2017}, the abstraction itself being learned by the algorithm avoiding information lost of statistical discretization.
% In \cite{TRILP2018}, a semantic free algorithm is proposed as a ways to model systems from raw data without burdening the modelers with an a priori choice of the proper semantics.
\citet{DMTRICAPS16} combine LFIT with a reinforcement learning algorithm to learn probabilistic models with exogenous effects (effects not related to any action) from scratch.
The learner was notably integrated with a robot to perform the task of clearing the tableware on a table.
In this task external agents interacted, people brought new tableware continuously and the manipulator robot had to cooperate with mobile robots to take the tableware to the kitchen.
The learner was able to learn a usable model in just five episodes of 30 action executions.
\citet{apperception} apply the \emph{Apperception Engine} to explain sequential data, such as cellular automata traces, rhythms and simple nursery tunes, image occlusion tasks,  game dynamics, and sequence induction intelligence tests.
Surprisingly, they show that their system can achieve human-level performance on the sequence induction intelligence tests in the zero-shot setting (without having been trained on lots of other examples of such tests, and without hand-engineered knowledge of the particular setting).
At a high level, these systems take the unique selling point of ILP systems (the ability to strongly generalise from a handful of data), and apply it to the self-supervised setting, producing an explicit human-readable theory that explains the observed state transitions.

% \paragraph{Other.}
% Other notable applications include learning event recognition systems ~\cite{iled,oled}, tracking the evolution of online communities~\cite{DBLP:conf/time/AthanasopoulosP18}, and the MNIST dataset \cite{dilp}.

\section{Summary and future work}
\label{sec:cons}

In a survey paper from a decade ago, \citet{ilp20} proposed directions for future research.
In the decade since, there have been major advances on many of the topics, notably in predicate invention (Section \ref{sec:pi}), using higher-order logic as a representation language (Section \ref{sec:metarules}) and to represent hypotheses (Section \ref{sec:ho}), and applications in learning actions and strategies (Section \ref{sec:robotics}).
Despite the advances, there are still many limitations in ILP that future work should address.

\subsection{Limitations and future research}

\paragraph{Better systems.}

\citet{ilp20} argue that a problem with ILP is the lack of well-engineered tools.
They state that whilst over 100 ILP systems have been built, less than a handful of systems can be meaningfully used by ILP researchers.
In the decade since the authors highlighted this problem, little progress has been made: most ILP systems are not easy to use.
In other words, ILP systems are still notoriously difficult to use and you often need a PhD in ILP to use any of the tools.
Even then, it is still often only the developers of a system that know how to properly use it.
By contrast, driven by industry, other forms of ML now have reliable and well-maintained implementations, such as PyTorch and TensorFlow, which has helped drive research.
A frustrating issue with ILP systems is that they use many different language biases or even different syntax for the same biases.
For instance, the way of specifying a learning task in Progol, Aleph, TILDE, and ILASP varies considerably despite them all using mode declarations.
If it is difficult for ILP researchers to use ILP tools, then what hope do non-ILP researchers have?
For ILP to be more widely adopted both inside and outside of academia, we must develop more standardised, user-friendly, and better-engineered tools.

\paragraph{Language biases.}
As \citet{ilp30} state, one major issue with ILP is choosing an appropriate language bias.
For instance, Metagol uses metarules (Section \ref{sec:metarules}) to restrict the syntax of hypotheses and thus the hypothesis space.
If a user can provide suitable metarules, then Metagol is extremely efficient.
However, if a user cannot provide suitable metarules (which is often the case), then Metagol is almost useless.
This same brittleness applies to ILP systems that employ mode declarations \cite{progol}.
In theory, a user can provide very general mode declarations, such as only using a single type and allowing unlimited recall.
In practice, however, weak mode declarations often lead to very poor performance.
For good performance, users of mode-based systems often need to manually analyse a given learning task to tweak the mode declarations, often through a process of trial and error.
Moreover, if a user makes a small mistake with a mode declaration, such as giving the wrong argument type, then the ILP system is unlikely to find a good solution.
Even for ILP experts, determining a suitable language bias is often a frustrating and time-consuming process.
We think the need for an almost perfect language bias is severely holding back ILP from being widely adopted.
By contrast, there are some neural net architectures (e.g. the transformer \cite{vaswani2017attention}) that can be applied successfully to a large range of diverse problems without requiring any domain-specific tuning.
We think that an important direction for future work in ILP is to develop techniques for automatically identifying suitable language biases.
Although there is some work on mode learning \cite{modelearning,DBLP:conf/ilp/FerilliEBM04,DBLP:conf/sigmod/PicadoTFP17,DBLP:conf/sigmod/PicadoTFPID21} and work on identifying suitable metarules \cite{reduce}, this area of research is largely under-researched.

\paragraph{Better datasets.}
Interesting problems, alongside usable systems, drive research and attract interest in a research field.
This relationship is most evident in the deep learning community which has, over a decade, grown into the largest AI community.
This community growth has been supported by the constant introduction of new problems, datasets, and well-engineered tools.
Challenging problems that push the state-of-the-art to its limits are essential to sustain progress in the field; otherwise, the field risks stagnation through only small incremental progress.
ILP has, unfortunately, failed to deliver on this front: most research is still evaluated on 20-year old datasets.
Most new datasets that have been introduced often come from toy domains and are designed to test specific properties of the introduced technique.
To an outsider, this sends a message that ILP is not applicable to real-world problems.
We think that the ILP community should learn from the experiences of other AI communities and put significant efforts into developing datasets that identify limitations of existing methods as well as showcase potential applications of ILP.
After all, it is no coincidence that SAT solving performance increased dramatically after the introduction of the SAT solving competitions \cite{jarvisalo2012international}.

\paragraph{Relevance.}
New methods for predicate invention (Section \ref{sec:pi}) have improved the abilities of ILP systems to invent high-level concepts.
These techniques raise the potential for ILP to be used in lifelong learning settings.
However, inventing and acquiring new BK could lead to a problem of too much BK, which can overwhelm an ILP system \cite{ashwin:badbk,forgetgol}.
On this issue, a key under-explored topic is that of \emph{relevancy}.
Given a new induction problem with large amounts of BK, how does an ILP system decide which BK is relevant?
One emerging technique is to train a neural network to score how relevant programs are in the BK and to then only use BK with the highest score to learn programs \cite{deepcoder,ellis:scc}.
However, the empirical efficacy of this approach has yet to be demonstrated.
Moreover, these approaches have only been demonstrated on small amounts of BK and it is unclear how they scale to BK with thousands of relations.
Without efficient methods of relevance identification, it is unclear how efficient lifelong learning can be achieved.

% \paragraph{Relevance.}
% \ac{plagarises the JAIR submission}
% The \emph{catastrophic remembering} problem is essentially the problem of \emph{relevance}: given a new ILP problem with lots of BK, how does an ILP system decide which BK is relevant?
% Although too much irrelevant BK is detrimental to learning performance \cite{ashwin:badbk}, there is almost no work in ILP on trying to identify relevant BK.
% One emerging technique is to train a neural network to score how relevant programs are in the BK and to then only use BK with the highest score to learn programs \cite{deepcoder,ellis:scc}.
% However, the empirical efficacy of this approach has yet to be demonstrated.
% Moreover, these approaches have only been demonstrated on small amounts of BK and it is unclear how they scale to BK with thousands of relations.
% Without efficient relevancy methods, it is unclear how lifelong learning can be achieved.

\paragraph{Handling mislabelled and ambiguous data.}
A major open question in ILP is how best to handle noisy and ambiguous data.
% Most symbolic ILP systems assume that the input examples are noise-free, although there are exceptions \cite{law:noisy}.
Neural ILP systems \cite{DBLP:conf/nips/Rocktaschel017,dilp} are designed from the start to robustly handle mislabelled data.
Although there has been work in recent years on designing ILP systems that can handle noisy mislabelled data, there is much less work on the even harder and more fundamental problem of designing ILP systems that can handle \emph{raw ambiguous data}.
ILP systems typically assume that the input has already been preprocessed into symbolic declarative form (typically, a set of ground atoms representing positive and negative examples).
But real-world input does not arrive in symbolic form.
Consider e.g. a robot with a video camera, where the raw input is a sequence of pixel images.
Converting each pixel image into a set of ground atoms is a challenging non-trivial achievement that should not be taken for granted.
For ILP systems to be widely applicable in the real world, they need to be redesigned so they can handle raw ambiguous input from the outset \cite{dilp,NLM}.

%\paragraph{Noisy BK.}
%Another issue related to lifelong learning is the underlying uncertainty associated with adding learned programs to the BK.
%By the inherent nature of induction, induced programs are not guaranteed to be correct (i.e. are expected to be noisy), yet they are the building blocks for subsequent induction.
%Building noisy programs on top of other noisy programs could lead to eventual incoherence of the learned program.
%This issue is especially problematic because most ILP approaches assume noiseless BK, i.e. a relation is true or false without any room for uncertainty.
%One of the appealing features of \dilp{} is that it takes a differentiable approach to ILP, where it can be provided with fuzzy or ambiguous data.
%Developing similar techniques to handle noisy BK is an under-explored topic in ILP.

\paragraph{Probabilistic ILP.}
Real-world data is often noisy and uncertain.
Extending ILP to deal with such uncertainty substantially broadens its applicability.
While StarAI is receiving growing attention, learning probabilistic programs from data is still largely under-investigated due to the complexity of joint probabilistic and logical inference.
When working with probabilistic programs, we are interested in the probability that a program covers an example, not only whether the program covers the example.
Consequently, probabilistic programs need to compute all possible derivations of an example, not just a single one.
Despite added complexity, probabilistic ILP opens many new challenges.
Most of the existing work on probabilistic ILP considers the minimal extension of ILP to the probabilistic setting, by assuming that either (i) BK facts are uncertain, or (ii) that learned clauses need to model uncertainty.
These assumptions make it possible to separate structure from uncertainty and simply reuse existing ILP techniques.
Following this minimal extension, the existing work focuses on discriminative learning in which the goal is to learn a program for a single target relation.
However, a grand challenge in probabilistic programming is generative learning.
That is, learning a program describing a generative process behind the data, not a single target relation.
Learning generative programs is a significantly more challenging problem, which has received very little attention in probabilistic ILP.

\paragraph{Explainability.}
Explainability is one of the claimed advantages of a symbolic representation. Recent work \cite{mugg:compmlj,mugg:beneficial} evaluates the comprehensibility of ILP hypotheses using Michie's \cite{usml} framework of \emph{ultra-strong machine learning}, where a learned hypothesis is expected to not only be accurate but to also demonstrably improve the performance of a human being provided with the learned hypothesis.
\cite{mugg:compmlj} empirically demonstrate improved human understanding directly through learned hypotheses. However, given the demonstration of both beneficial and harmful effects of explainability \cite{mugg:beneficial} more work is required to better understand the conditions under which this can be achieved, especially given the rise of PI.

\paragraph{Unifying ILP with neural methods.}
It has often been noted \cite{dilp} that the strengths and weaknesses of neural networks and ILP are \emph{complementary}: neural networks (1) scale to huge datasets, (2) are robust to mislabelled data, (3) are robust to ambiguous (raw, undiscretised) data, but (4) are very data hungry, (5) often struggle to generalise outside the training distribution, and (6) are uninterpretable. ILP systems, by contrast (1) often fail to scale to large datasets, (2) sometimes fail to handle mislabelled data, (3) almost always fail to handle raw undiscretised data, but (4) are very data efficient, (5) often generalise well outside the training distribution, and (6) produce human-readable programs. Given that the strengths and weaknesses of the two approaches are complementary, many people have advocated some sort of unification of the two \cite{DeRaedtKerstingEtAl16,garcez2020neurosymbolic,dilp}. There is much activity in this area, and much work still to do to produce a truly convincing unification of these very different paradigms.

\subsection{Summary}
% As ILP approaches 30, we think that the recent advances surveyed in this paper put ILP in a prime position to have a significant impact on AI over the next decade especially to address the main limitations of state-of-the-art machine learning.
As ILP approaches 30, we think that the advances made in the last decade, surveyed in this paper, have opened up new areas of research for ILP to explore.
Moreover, we hope that the next decade sees developments on the numerous limitations we have discussed so that ILP can have a significant impact on AI.

% "As ILP approaches 30, we think that the recent advances surveyed in this paper have opened up new areas of research for ILP to explore in the next decade. Moreover, we hope that the next decade sees developments not eh numerous limitations we have discussed, so that ILP can have a significant impact on AI over the next decade.”

% We think that these and other recent advances put ILP in a prime position to have a significant impact on AI over the next decade, especially to address the key limitations of standard forms of machine learning.

\section*{Declarations}

\paragraph{Funding.}
Not applicable.

\paragraph{Conflicts of interest/Competing interests.}
Not applicable.

\paragraph{Ethics approval.}
Not applicable.

\paragraph{Consent to participate.}
Not applicable.

\paragraph{Consent for publication.}
Not applicable.

\paragraph{Availability of data and material.}
Not applicable.

\paragraph{Code availability.}
Not applicable.

\paragraph{Consent to participate.}
Not applicable.

\paragraph{Authors' contributions.}
AC wrote Sections 1-7, and 9.
SD wrote Sections 4.3, 5.4, and 8.
RE wrote Sections 7.2, 8, and 9.
SM wrote Section 9.

\bibliographystyle{spbasic}
\bibliography{ourbib}

\begin{thebibliography}{117}
\providecommand{\natexlab}[1]{#1}
\providecommand{\url}[1]{{#1}}
\providecommand{\urlprefix}{URL }
\expandafter\ifx\csname urlstyle\endcsname\relax
  \providecommand{\doi}[1]{DOI~\discretionary{}{}{}#1}\else
  \providecommand{\doi}{DOI~\discretionary{}{}{}\begingroup
  \urlstyle{rm}\Url}\fi
\providecommand{\eprint}[2][]{\url{#2}}

\bibitem[{Ahlgren and Yuen(2013)}]{atom}
Ahlgren J, Yuen SY (2013) Efficient program synthesis using constraint
  satisfaction in inductive logic programming. J Machine Learning Res
  14(1):3649--3682

\bibitem[{Ai et~al(2020)Ai, Muggleton, Hocquette, Gromowski, and
  Schmid}]{mugg:beneficial}
Ai L, Muggleton S, Hocquette C, Gromowski M, Schmid U (2020) Beneficial and
  harmful explanatory machine learning. Machine Learning In Press, available
  http://arxiv.org/abs/2009.06410

\bibitem[{Albarghouthi et~al(2017)Albarghouthi, Koutris, Naik, and
  Smith}]{awscp}
Albarghouthi A, Koutris P, Naik M, Smith C (2017) Constraint-based synthesis of
  datalog programs. In: Principles and Practice of Constraint Programming -
  23rd International Conference, {CP} 2017, Springer, Lecture Notes in Computer
  Science, vol 10416, pp 689--706

\bibitem[{Antanas et~al(2015)Antanas, Moreno, and {De Raedt}}]{AntanasMR15}
Antanas L, Moreno P, {De Raedt} L (2015) Relational kernel-based grasping with
  numerical features. In: Inductive Logic Programming - 25th International
  Conference, {ILP} 2015, Springer, Lecture Notes in Computer Science, vol
  9575, pp 1--14

\bibitem[{Bain and Muggleton(1992)}]{Bain92non-monotoniclearning}
Bain M, Muggleton S (1992) Non-monotonic learning. In: Inductive Logic
  Programming, Academic Press, pp 145--161

\bibitem[{Bain and Srinivasan(2018)}]{DBLP:journals/ml/BainS18}
Bain M, Srinivasan A (2018) Identification of biological transition systems
  using meta-interpreted logic programs. Machine Learning 107(7):1171--1206

\bibitem[{Balog et~al(2017)Balog, Gaunt, Brockschmidt, Nowozin, and
  Tarlow}]{deepcoder}
Balog M, Gaunt AL, Brockschmidt M, Nowozin S, Tarlow D (2017) Deepcoder:
  Learning to write programs. In: 5th International Conference on Learning
  Representations, {ICLR} 2017, OpenReview.net

\bibitem[{Bartha and Cheney(2019)}]{DBLP:conf/ilp/BarthaC19}
Bartha S, Cheney J (2019) Towards meta-interpretive learning of programming
  language semantics. In: Inductive Logic Programming - 29th International
  Conference, {ILP} 2019, Springer, Lecture Notes in Computer Science, vol
  11770, pp 16--25

\bibitem[{Bellodi and Riguzzi(2015)}]{DBLP:journals/tplp/BellodiR15}
Bellodi E, Riguzzi F (2015) Structure learning of probabilistic logic programs
  by searching the clause space. Theory Pract Log Program 15(2):169--212

\bibitem[{Blockeel and {De Raedt}(1998)}]{tilde}
Blockeel H, {De Raedt} L (1998) Top-down induction of first-order logical
  decision trees. Artif Intell 101(1-2):285--297

\bibitem[{Bohan et~al(2011)Bohan, Caron-Lormier, Muggleton, Raybould, and
  Tamaddoni-Nezhad}]{bohan2011automated}
Bohan DA, Caron-Lormier G, Muggleton S, Raybould A, Tamaddoni-Nezhad A (2011)
  Automated discovery of food webs from ecological data using logic-based
  machine learning. PLoS One 6(12):e29,028

\bibitem[{Bohan et~al(2017)Bohan, Vacher, Tamaddoni-Nezhad, Raybould, Dumbrell,
  and Woodward}]{bohan2017next}
Bohan DA, Vacher C, Tamaddoni-Nezhad A, Raybould A, Dumbrell AJ, Woodward G
  (2017) Next-generation global biomonitoring: large-scale, automated
  reconstruction of ecological networks. Trends in Ecology \& Evolution
  32(7):477--487

\bibitem[{Bratko(1999)}]{hyper}
Bratko I (1999) Refining complete hypotheses in {ILP}. In: Inductive Logic
  Programming, 9th International Workshop, ILP-99, Springer, Lecture Notes in
  Computer Science, vol 1634, pp 44--55

\bibitem[{Chollet(2019)}]{chollet:2019}
Chollet F (2019) On the measure of intelligence. CoRR abs/1911.01547

\bibitem[{Corapi et~al(2011)Corapi, Russo, and Lupu}]{aspal}
Corapi D, Russo A, Lupu E (2011) Inductive logic programming in answer set
  programming. In: Inductive Logic Programming - 21st International Conference,
  {ILP} 2011, Springer, Lecture Notes in Computer Science, vol 7207, pp 91--97

\bibitem[{Cropper(2019)}]{playgol}
Cropper A (2019) Playgol: Learning programs through play. In: Proceedings of
  the Twenty-Eighth International Joint Conference on Artificial Intelligence,
  {IJCAI} 2019, ijcai.org, pp 6074--6080

\bibitem[{Cropper(2020)}]{forgetgol}
Cropper A (2020) Forgetting to learn logic programs. In: The Thirty-Fourth
  {AAAI} Conference on Artificial Intelligence, {AAAI} 2020, {AAAI} Press, pp
  3676--3683

\bibitem[{Cropper and Dumancic(2020)}]{ilpintro}
Cropper A, Dumancic S (2020) Inductive logic programming at 30: a new
  introduction. CoRR abs/2008.07912,
  \urlprefix\url{https://arxiv.org/abs/2008.07912}, \eprint{2008.07912}

\bibitem[{Cropper and Dumančić(2020)}]{brute}
Cropper A, Dumančić S (2020) Learning large logic programs by going beyond
  entailment. In: Proceedings of the Twenty-Ninth International Joint
  Conference on Artificial Intelligence, {IJCAI} 2020, ijcai.org, pp 2073--2079

\bibitem[{Cropper and Morel(2021{\natexlab{a}})}]{popper}
Cropper A, Morel R (2021{\natexlab{a}}) Learning programs by learning from
  failures. Machine Learning 110(4):801--856, \doi{10.1007/s10994-020-05934-z},
  \urlprefix\url{https://doi.org/10.1007/s10994-020-05934-z}

\bibitem[{Cropper and Morel(2021{\natexlab{b}})}]{poppi}
Cropper A, Morel R (2021{\natexlab{b}}) Predicate invention by learning from
  failures. CoRR abs/2104.14426,
  \urlprefix\url{https://arxiv.org/abs/2104.14426}, \eprint{2104.14426}

\bibitem[{Cropper and Muggleton(2015)}]{metagolo}
Cropper A, Muggleton SH (2015) Learning efficient logical robot strategies
  involving composable objects. In: Proceedings of the Twenty-Fourth
  International Joint Conference on Artificial Intelligence, {IJCAI} 2015,
  {AAAI} Press, pp 3423--3429

\bibitem[{Cropper and Muggleton(2016)}]{metagol}
Cropper A, Muggleton SH (2016) Metagol system.
  \urlprefix\url{https://github.com/metagol/metagol}

\bibitem[{Cropper and Muggleton(2019)}]{metaopt}
Cropper A, Muggleton SH (2019) Learning efficient logic programs. Machine
  Learning 108(7):1063--1083

\bibitem[{Cropper and Tourret(2020)}]{reduce}
Cropper A, Tourret S (2020) Logical reduction of metarules. Machine Learning
  109(7):1323--1369

\bibitem[{Cropper et~al(2015)Cropper, Tamaddoni{-}Nezhad, and
  Muggleton}]{datacurate}
Cropper A, Tamaddoni{-}Nezhad A, Muggleton SH (2015) Meta-interpretive learning
  of data transformation programs. In: Inductive Logic Programming - 25th
  International Conference, {ILP} 2015, Springer, Lecture Notes in Computer
  Science, vol 9575, pp 46--59

\bibitem[{Cropper et~al(2020{\natexlab{a}})Cropper, Dumančić, and
  Muggleton}]{ilp30}
Cropper A, Dumančić S, Muggleton SH (2020{\natexlab{a}}) Turning 30: New
  ideas in inductive logic programming. In: Proceedings of the Twenty-Ninth
  International Joint Conference on Artificial Intelligence, {IJCAI} 2020,
  ijcai.org, pp 4833--4839

\bibitem[{Cropper et~al(2020{\natexlab{b}})Cropper, Evans, and Law}]{iggp}
Cropper A, Evans R, Law M (2020{\natexlab{b}}) Inductive general game playing.
  Machine Learning 109(7):1393–1434

\bibitem[{Cropper et~al(2020{\natexlab{c}})Cropper, Morel, and
  Muggleton}]{metaho}
Cropper A, Morel R, Muggleton S (2020{\natexlab{c}}) Learning higher-order
  logic programs. Machine Learning 109(7):1289–1322

\bibitem[{Dai and Muggleton(2021)}]{muggzhou:neuralmil}
Dai WZ, Muggleton SH (2021) Abductive knowledge induction from raw data. In:
  Proceedings of the 35th Conference on Artificial Intelligence (IJCAI 2021),
  IJCAI, in Press

\bibitem[{{De Raedt}(2008)}]{luc:book}
{De Raedt} L (2008) Logical and relational learning. Cognitive Technologies,
  Springer

\bibitem[{{De Raedt} and Kersting(2008)}]{pilp}
{De Raedt} L, Kersting K (2008) Probabilistic Inductive Logic Programming,
  Springer-Verlag, Berlin, Heidelberg, p 1–27

\bibitem[{{De Raedt} et~al(2007){De Raedt}, Kimmig, and
  Toivonen}]{raedt:problog}
{De Raedt} L, Kimmig A, Toivonen H (2007) Problog: {A} probabilistic prolog and
  its application in link discovery. In: {IJCAI} 2007, Proceedings of the 20th
  International Joint Conference on Artificial Intelligence, Hyderabad, India,
  January 6-12, 2007, pp 2462--2467

\bibitem[{{De Raedt} et~al(2015){De Raedt}, Dries, Thon, den Broeck, and
  Verbeke}]{probfoil}
{De Raedt} L, Dries A, Thon I, den Broeck GV, Verbeke M (2015) Inducing
  probabilistic relational rules from probabilistic examples. In: Proceedings
  of the Twenty-Fourth International Joint Conference on Artificial
  Intelligence, {IJCAI} 2015, {AAAI} Press, pp 1835--1843

\bibitem[{{De Raedt} et~al(2016){De Raedt}, Kersting, Natarajan, and
  Poole}]{DeRaedtKerstingEtAl16}
{De Raedt} L, Kersting K, Natarajan S, Poole D (2016) Statistical Relational
  Artificial Intelligence: Logic, Probability, and Computation. Synthesis
  Lectures on Artificial Intelligence and Machine Learning, Morgan {\&}
  Claypool Publishers

\bibitem[{Dong et~al(2019)Dong, Mao, Lin, Wang, Li, and Zhou}]{NLM}
Dong H, Mao J, Lin T, Wang C, Li L, Zhou D (2019) Neural logic machines. In:
  7th International Conference on Learning Representations, {ICLR} 2019, New
  Orleans, LA, USA, May 6-9, 2019, OpenReview.net,
  \urlprefix\url{https://openreview.net/forum?id=B1xY-hRctX}

\bibitem[{Dumancic et~al(2020)Dumancic, Guns, and Cropper}]{knorf}
Dumancic S, Guns T, Cropper A (2020) Knowledge refactoring for inductive
  program synthesis. {AAAI}

\bibitem[{Dumančić and Blockeel(2017)}]{curled}
Dumančić S, Blockeel H (2017) Clustering-based relational unsupervised
  representation learning with an explicit distributed representation. In:
  Proceedings of the Twenty-Sixth International Joint Conference on Artificial
  Intelligence, {IJCAI} 2017, ijcai.org, pp 1631--1637

\bibitem[{Dumančić et~al(2019)Dumančić, Guns, Meert, and Blockeel}]{alps}
Dumančić S, Guns T, Meert W, Blockeel H (2019) Learning relational
  representations with auto-encoding logic programs. In: Proceedings of the
  Twenty-Eighth International Joint Conference on Artificial Intelligence,
  {IJCAI} 2019, ijcai.org, pp 6081--6087

\bibitem[{Ellis et~al(2018)Ellis, Morales, Sabl{\'{e}}{-}Meyer, Solar{-}Lezama,
  and Tenenbaum}]{ellis:scc}
Ellis K, Morales L, Sabl{\'{e}}{-}Meyer M, Solar{-}Lezama A, Tenenbaum J (2018)
  Learning libraries of subroutines for neurally-guided bayesian program
  induction. In: NeurIPS 2018, pp 7816--7826

\bibitem[{Evans and Grefenstette(2018)}]{dilp}
Evans R, Grefenstette E (2018) Learning explanatory rules from noisy data. J
  Artif Intell Res 61:1--64

\bibitem[{Evans et~al(2021)Evans, Hernández-Orallo, Welbl, Kohli, and
  Sergot}]{apperception}
Evans R, Hernández-Orallo J, Welbl J, Kohli P, Sergot M (2021) Making sense of
  sensory input. Artificial Intelligence p 103438

\bibitem[{Ferilli(2016)}]{DBLP:journals/jiis/Ferilli16a}
Ferilli S (2016) Predicate invention-based specialization in inductive logic
  programming. J Intell Inf Syst \doi{10.1007/s10844-016-0412-9},
  \urlprefix\url{https://doi.org/10.1007/s10844-016-0412-9}

\bibitem[{Ferilli et~al(2004)Ferilli, Esposito, Basile, and
  Mauro}]{DBLP:conf/ilp/FerilliEBM04}
Ferilli S, Esposito F, Basile TMA, Mauro ND (2004) Automatic induction of
  first-order logic descriptors type domains from observations. In: Inductive
  Logic Programming, 14th International Conference, {ILP} 2004, Springer,
  Lecture Notes in Computer Science, vol 3194, pp 116--131

\bibitem[{Garcez and Lamb(2020)}]{garcez2020neurosymbolic}
Garcez Ad, Lamb LC (2020) Neurosymbolic ai: the 3rd wave. arXiv preprint
  arXiv:201205876

\bibitem[{Gebser et~al(2012{\natexlab{a}})Gebser, Kaminski, Kaufmann, and
  Schaub}]{asp}
Gebser M, Kaminski R, Kaufmann B, Schaub T (2012{\natexlab{a}}) Answer Set
  Solving in Practice. Synthesis Lectures on Artificial Intelligence and
  Machine Learning, Morgan {\&} Claypool Publishers

\bibitem[{Gebser et~al(2012{\natexlab{b}})Gebser, Kaufmann, and Schaub}]{clasp}
Gebser M, Kaufmann B, Schaub T (2012{\natexlab{b}}) Conflict-driven answer set
  solving: From theory to practice. Artif Intell 187:52--89

\bibitem[{Genesereth and Bj{\"{o}}rnsson(2013)}]{ggp}
Genesereth MR, Bj{\"{o}}rnsson Y (2013) The international general game playing
  competition. {AI} Magazine 34(2):107--111

\bibitem[{Gulwani(2011)}]{flashfill}
Gulwani S (2011) Automating string processing in spreadsheets using
  input-output examples. In: Proceedings of the 38th {ACM} {SIGPLAN-SIGACT}
  Symposium on Principles of Programming Languages, {POPL} 2011, {ACM}, pp
  317--330

\bibitem[{Heule et~al(2016)Heule, Kullmann, and Marek}]{satprogresss}
Heule MJH, Kullmann O, Marek VW (2016) Solving and verifying the boolean
  pythagorean triples problem via cube-and-conquer. In: Creignou N, Berre DL
  (eds) Theory and Applications of Satisfiability Testing - {SAT} 2016 - 19th
  International Conference, Bordeaux, France, July 5-8, 2016, Proceedings,
  Springer, Lecture Notes in Computer Science, vol 9710, pp 228--245,
  \doi{10.1007/978-3-319-40970-2\_15},
  \urlprefix\url{https://doi.org/10.1007/978-3-319-40970-2\_15}

\bibitem[{Hocquette and Muggleton(2020)}]{celine:bottom}
Hocquette C, Muggleton SH (2020) Complete bottom-up predicate invention in
  meta-interpretive learning. In: Proceedings of the Twenty-Ninth International
  Joint Conference on Artificial Intelligence, {IJCAI} 2020, ijcai.org, pp
  2312--2318

\bibitem[{Huynh and Mooney(2008)}]{huynh:icml08}
Huynh TN, Mooney RJ (2008) Discriminative structure and parameter learning for
  markov logic networks. In: Proceedings of the 25th International Conference
  on Machine Learning, Association for Computing Machinery, New York, NY, USA,
  p 416?423, \doi{10.1145/1390156.1390209}

\bibitem[{Inoue(2016)}]{inoue:flap}
Inoue K (2016) Meta-level abduction. {FLAP} 3(1):7--36

\bibitem[{Inoue et~al(2013)Inoue, Doncescu, and Nabeshima}]{inoue:mla}
Inoue K, Doncescu A, Nabeshima H (2013) Completing causal networks by
  meta-level abduction. Machine Learning 91(2):239--277

\bibitem[{Inoue et~al(2014)Inoue, Ribeiro, and Sakama}]{inoue:lfit}
Inoue K, Ribeiro T, Sakama C (2014) Learning from interpretation transition.
  Machine Learning 94(1):51--79

\bibitem[{J{\"a}rvisalo et~al(2012)J{\"a}rvisalo, Le~Berre, Roussel, and
  Simon}]{jarvisalo2012international}
J{\"a}rvisalo M, Le~Berre D, Roussel O, Simon L (2012) The international sat
  solver competitions. Ai Magazine 33(1):89--92

\bibitem[{Kaalia et~al(2016)Kaalia, Srinivasan, Kumar, and Ghosh}]{Kaalia16}
Kaalia R, Srinivasan A, Kumar A, Ghosh I (2016) {ILP}-assisted de novo drug
  design. Machine Learning 103(3):309--341

\bibitem[{Kaiser and Sutskever(2016)}]{DBLP:journals/corr/KaiserS15}
Kaiser L, Sutskever I (2016) Neural gpus learn algorithms. In: 4th
  International Conference on Learning Representations, {ICLR} 2016

\bibitem[{Kaminski et~al(2018)Kaminski, Eiter, and Inoue}]{hexmil}
Kaminski T, Eiter T, Inoue K (2018) Exploiting answer set programming with
  external sources for meta-interpretive learning. Theory Pract Log Program
  18(3-4):571--588

\bibitem[{Katzouris et~al(2015)Katzouris, Artikis, and Paliouras}]{iled}
Katzouris N, Artikis A, Paliouras G (2015) Incremental learning of event
  definitions with inductive logic programming. Machine Learning
  100(2-3):555--585

\bibitem[{Katzouris et~al(2016)Katzouris, Artikis, and Paliouras}]{oled}
Katzouris N, Artikis A, Paliouras G (2016) Online learning of event
  definitions. Theory Pract Log Program 16(5-6):817--833

\bibitem[{Kok and Domingos(2009)}]{kok:icml09}
Kok S, Domingos P (2009) Learning markov logic network structure via hypergraph
  lifting. In: Proceedings of the 26th International Conference on Machine
  Learning, Association for Computing Machinery, New York, NY, USA, p 505?512,
  \doi{10.1145/1553374.1553440}

\bibitem[{Kok and Domingos(2007)}]{pedro:pi}
Kok S, Domingos PM (2007) Statistical predicate invention. In: Machine
  Learning, Proceedings of the Twenty-Fourth International Conference {(ICML}
  2007), {ACM}, {ACM} International Conference Proceeding Series, vol 227, pp
  433--440

\bibitem[{Kramer(1995)}]{kramer1995predicate}
Kramer S (1995) Predicate invention: A comprehensive view. Rapport technique
  OFAI-TR-95-32, Austrian Research Institute for Artificial Intelligence,
  Vienna

\bibitem[{Kramer(2020)}]{kramer:ijcai20}
Kramer S (2020) A brief history of learning symbolic higher-level
  representations from data (and a curious look forward). In: Proceedings of
  the Twenty-Ninth International Joint Conference on Artificial Intelligence,
  {IJCAI} 2020, ijcai.org, pp 4868--4876

\bibitem[{Law(2018)}]{ilasp3}
Law M (2018) Inductive learning of answer set programs. PhD thesis, Imperial
  College London, {UK}

\bibitem[{Law et~al(2014)Law, Russo, and Broda}]{ilasp}
Law M, Russo A, Broda K (2014) Inductive learning of answer set programs. In:
  Logics in Artificial Intelligence - 14th European Conference, {JELIA} 2014,
  Springer, Lecture Notes in Computer Science, vol 8761, pp 311--325

\bibitem[{Law et~al(2018)Law, Russo, and Broda}]{law:aij}
Law M, Russo A, Broda K (2018) The complexity and generality of learning answer
  set programs. Artif Intell 259:110--146

\bibitem[{Law et~al(2019)Law, Russo, Bertino, Broda, and Lobo}]{law:asg}
Law M, Russo A, Bertino E, Broda K, Lobo J (2019) Representing and learning
  grammars in answer set programming. In: The Thirty-Third {AAAI} Conference on
  Artificial Intelligence, {AAAI} 2019, {AAAI} Press, pp 2919--2928

\bibitem[{Law et~al(2020{\natexlab{a}})Law, Russo, Bertino, Broda, and
  Lobo}]{law:fastlas}
Law M, Russo A, Bertino E, Broda K, Lobo J (2020{\natexlab{a}}) Fastlas:
  Scalable inductive logic programming incorporating domain-specific
  optimisation criteria. In: The Thirty-Fourth {AAAI} Conference on Artificial
  Intelligence, {AAAI} 2020, {AAAI} Press, pp 2877--2885

\bibitem[{Law et~al(2020{\natexlab{b}})Law, Russo, and Broda}]{law:alp}
Law M, Russo A, Broda K (2020{\natexlab{b}}) The ilasp system for inductive
  learning of answer set programs. The Association for Logic Programming
  Newsletter

\bibitem[{Leban et~al(2008)Leban, Zabkar, and Bratko}]{DBLP:conf/ilp/LebanZB08}
Leban G, Zabkar J, Bratko I (2008) An experiment in robot discovery with {ILP}.
  In: Inductive Logic Programming, 18th International Conference, {ILP} 2008,
  Springer, Lecture Notes in Computer Science, vol 5194, pp 77--90

\bibitem[{Legras et~al(2018)Legras, Rouveirol, and
  Ventos}]{DBLP:conf/ilp/LegrasRV18}
Legras S, Rouveirol C, Ventos V (2018) The game of bridge: {A} challenge for
  {ILP}. In: Inductive Logic Programming - 28th International Conference, {ILP}
  2018, Springer, Lecture Notes in Computer Science, vol 11105, pp 72--87

\bibitem[{Lin et~al(2014)Lin, Dechter, Ellis, Tenenbaum, and
  Muggleton}]{metabias}
Lin D, Dechter E, Ellis K, Tenenbaum JB, Muggleton S (2014) Bias reformulation
  for one-shot function induction. In: {ECAI} 2014 - 21st European Conference
  on Artificial Intelligence, 18-22 August 2014, {IOS} Press, Frontiers in
  Artificial Intelligence and Applications, vol 263, pp 525--530

\bibitem[{Marcus(2018)}]{marcus:2018}
Marcus G (2018) Deep learning: {A} critical appraisal. CoRR abs/1801.00631

\bibitem[{Mart{\'{\i}}nez et~al(2016)Mart{\'{\i}}nez, Aleny{\`{a}}, Torras,
  Ribeiro, and Inoue}]{DMTRICAPS16}
Mart{\'{\i}}nez D, Aleny{\`{a}} G, Torras C, Ribeiro T, Inoue K (2016) Learning
  relational dynamics of stochastic domains for planning. In: Proceedings of
  the Twenty-Sixth International Conference on Automated Planning and
  Scheduling, {ICAPS} 2016, {AAAI} Press, pp 235--243

\bibitem[{McCreath and Sharma(1995)}]{modelearning}
McCreath E, Sharma A (1995) Extraction of meta-knowledge to restrict the
  hypothesis space for ilp systems. In: Eighth Australian Joint Conference on
  Artificial Intelligence, pp 75--82

\bibitem[{Michie(1988)}]{usml}
Michie D (1988) Machine learning in the next five years. In: Sleeman DH (ed)
  Proceedings of the Third European Working Session on Learning, {EWSL} 1988,
  Turing Institute, Pitman Publishing, pp 107--122

\bibitem[{Muggleton(1987)}]{duce}
Muggleton S (1987) Duce, an oracle-based approach to constructive induction.
  In: Proceedings of the 10th International Joint Conference on Artificial
  Intelligence., Morgan Kaufmann, pp 287--292

\bibitem[{Muggleton(1991)}]{mugg:ilp}
Muggleton S (1991) Inductive logic programming. New Generation Computing
  8(4):295--318

\bibitem[{Muggleton(1995)}]{progol}
Muggleton S (1995) Inverse entailment and progol. New Generation Comput
  13(3{\&}4):245--286

\bibitem[{Muggleton and Buntine(1988)}]{cigol}
Muggleton S, Buntine WL (1988) Machine invention of first order predicates by
  inverting resolution. In: Machine Learning, Proceedings of the Fifth
  International Conference on Machine Learning, Morgan Kaufmann, pp 339--352

\bibitem[{Muggleton and {De Raedt}(1994)}]{mugg:ilp94}
Muggleton S, {De Raedt} L (1994) Inductive logic programming: Theory and
  methods. J Log Program 19/20:629--679

\bibitem[{Muggleton and Feng(1990)}]{golem}
Muggleton S, Feng C (1990) Efficient induction of logic programs. In:
  Algorithmic Learning Theory, First International Workshop, {ALT} '90, pp
  368--381

\bibitem[{Muggleton and Hocquette(2019)}]{mugghoq:newgengamestrat}
Muggleton S, Hocquette C (2019) Machine discovery of comprehensible strategies
  for simple games using meta-interpretive learning. New Generation Computing
  37:203--217

\bibitem[{Muggleton et~al(2009)Muggleton, Paes, Costa, and
  Zaverucha}]{chess-revision}
Muggleton S, Paes A, Costa VS, Zaverucha G (2009) Chess revision: Acquiring the
  rules of chess variants through {FOL} theory revision from examples. In:
  Inductive Logic Programming, 19th International Conference, {ILP} 2009,
  Springer, Lecture Notes in Computer Science, vol 5989, pp 123--130

\bibitem[{Muggleton et~al(2012)Muggleton, {De Raedt}, Poole, Bratko, Flach,
  Inoue, and Srinivasan}]{ilp20}
Muggleton S, {De Raedt} L, Poole D, Bratko I, Flach PA, Inoue K, Srinivasan A
  (2012) {ILP} turns 20 - biography and future challenges. Machine Learning
  86(1):3--23

\bibitem[{Muggleton et~al(2018{\natexlab{a}})Muggleton, Dai, Sammut,
  Tamaddoni{-}Nezhad, Wen, and Zhou}]{mugg:vision}
Muggleton S, Dai W, Sammut C, Tamaddoni{-}Nezhad A, Wen J, Zhou Z
  (2018{\natexlab{a}}) Meta-interpretive learning from noisy images. Machine
  Learning 107(7):1097--1118

\bibitem[{Muggleton et~al(2014)Muggleton, Lin, Pahlavi, and
  Tamaddoni{-}Nezhad}]{mugg:metalearn}
Muggleton SH, Lin D, Pahlavi N, Tamaddoni{-}Nezhad A (2014) Meta-interpretive
  learning: application to grammatical inference. Machine Learning 94(1):25--49

\bibitem[{Muggleton et~al(2015)Muggleton, Lin, and
  Tamaddoni{-}Nezhad}]{mugg:metagold}
Muggleton SH, Lin D, Tamaddoni{-}Nezhad A (2015) Meta-interpretive learning of
  higher-order dyadic {Datalog}: predicate invention revisited. Machine
  Learning 100(1):49--73

\bibitem[{Muggleton et~al(2018{\natexlab{b}})Muggleton, Schmid, Zeller,
  Tamaddoni{-}Nezhad, and Besold}]{mugg:compmlj}
Muggleton SH, Schmid U, Zeller C, Tamaddoni{-}Nezhad A, Besold TR
  (2018{\natexlab{b}}) Ultra-strong machine learning: comprehensibility of
  programs learned with {ILP}. Machine Learning 107(7):1119--1140

\bibitem[{Nienhuys-Cheng and Wolf(1997)}]{ilp:book}
Nienhuys-Cheng SH, Wolf Rd (1997) Foundations of Inductive Logic Programming.
  Springer-Verlag New York, Inc., Secaucus, NJ, USA

\bibitem[{Patsantzis and Muggleton(2021)}]{patmug:topprog}
Patsantzis S, Muggleton S (2021) Top program construction and reduction for
  polynomial time meta-interpretive learning. Machine Learning 110:755--778

\bibitem[{Picado et~al(2017)Picado, Termehchy, Fern, and
  Pathak}]{DBLP:conf/sigmod/PicadoTFP17}
Picado J, Termehchy A, Fern A, Pathak S (2017) Towards automatically setting
  language bias in relational learning. In: Proceedings of the 1st Workshop on
  Data Management for End-to-End Machine Learning, DEEM@SIGMOD 2017, {ACM}, pp
  3:1--3:4

\bibitem[{Picado et~al(2021)Picado, Termehchy, Fern, Pathak, Ilango, and
  Davis}]{DBLP:conf/sigmod/PicadoTFPID21}
Picado J, Termehchy A, Fern A, Pathak S, Ilango P, Davis J (2021) Scalable and
  usable relational learning with automatic language bias. In: Li G, Li Z,
  Idreos S, Srivastava D (eds) {SIGMOD} '21: International Conference on
  Management of Data, Virtual Event, China, June 20-25, 2021, {ACM}, pp
  1440--1451, \doi{10.1145/3448016.3457275},
  \urlprefix\url{https://doi.org/10.1145/3448016.3457275}

\bibitem[{Plotkin(1971)}]{plotkin:thesis}
Plotkin G (1971) Automatic methods of inductive inference. PhD thesis,
  Edinburgh University

\bibitem[{Quinlan(1990)}]{foil}
Quinlan JR (1990) Learning logical definitions from relations. Machine Learning
  5:239--266

\bibitem[{Ray(2009)}]{xhail}
Ray O (2009) Nonmonotonic abductive inductive learning. J Applied Logic
  7(3):329--340

\bibitem[{Reed and de~Freitas(2016)}]{nandopoo}
Reed SE, de~Freitas N (2016) Neural programmer-interpreters. In: 4th
  International Conference on Learning Representations, {ICLR} 2016

\bibitem[{Ribeiro and Inoue(2014)}]{DBLP:conf/ilp/RibeiroI14}
Ribeiro T, Inoue K (2014) Learning prime implicant conditions from
  interpretation transition. In: Inductive Logic Programming - 24th
  International Conference, {ILP} 2014, Springer, Lecture Notes in Computer
  Science, vol 9046, pp 108--125

\bibitem[{Ribeiro et~al(2015)Ribeiro, Magnin, Inoue, and Sakama}]{TRICMLA15}
Ribeiro T, Magnin M, Inoue K, Sakama C (2015) Learning multi-valued biological
  models with delayed influence from time-series observations. In: 14th {IEEE}
  International Conference on Machine Learning and Applications, {ICMLA} 2015,
  {IEEE}, pp 25--31

\bibitem[{Ribeiro et~al(2020)Ribeiro, Folschette, Magnin, and
  Inoue}]{TRMLJ2020}
Ribeiro T, Folschette M, Magnin M, Inoue K (2020) {Learning any semantics for
  dynamical systems represented by logic programs}, working paper or preprint

\bibitem[{Richardson and Domingos(2006)}]{richardson:mln}
Richardson M, Domingos PM (2006) Markov logic networks. Machine Learning
  62(1-2):107--136, \doi{10.1007/s10994-006-5833-1},
  \urlprefix\url{https://doi.org/10.1007/s10994-006-5833-1}

\bibitem[{Rockt{\"{a}}schel and Riedel(2017)}]{DBLP:conf/nips/Rocktaschel017}
Rockt{\"{a}}schel T, Riedel S (2017) End-to-end differentiable proving. In:
  Advances in Neural Information Processing Systems 30: Annual Conference on
  Neural Information Processing Systems 2017, 4-9 December 2017, pp 3788--3800

\bibitem[{Sammut et~al(2015)Sammut, Sheh, Haber, and Wicaksono}]{SammutSHW15}
Sammut C, Sheh R, Haber A, Wicaksono H (2015) The robot engineer. In: Late
  Breaking Papers of the 25th International Conference on Inductive Logic
  Programming, CEUR-WS.org, {CEUR} Workshop Proceedings, vol 1636, pp 101--106

\bibitem[{Sato(1995)}]{Sato95astatistical}
Sato T (1995) A statistical learning method for logic programs with
  distribution semantics. In: Sterling L (ed) Logic Programming, Proceedings of
  the Twelfth International Conference on Logic Programming, Tokyo, Japan, June
  13-16, 1995, {MIT} Press, pp 715--729

\bibitem[{Sato and Kameya(2001)}]{Sato2001ParameterLO}
Sato T, Kameya Y (2001) Parameter learning of logic programs for
  symbolic-statistical modeling. J Artif Intell Res 15:391--454,
  \doi{10.1613/jair.912}, \urlprefix\url{https://doi.org/10.1613/jair.912}

\bibitem[{Sch{\"{u}}ller and Benz(2018)}]{inspire}
Sch{\"{u}}ller P, Benz M (2018) Best-effort inductive logic programming via
  fine-grained cost-based hypothesis generation - the inspire system at the
  inductive logic programming competition. Machine Learning 107(7):1141--1169

\bibitem[{Sivaraman et~al(2019)Sivaraman, Zhang, den Broeck, and
  Kim}]{Sivaraman2019}
Sivaraman A, Zhang T, den Broeck GV, Kim M (2019) Active inductive logic
  programming for code search. In: Proceedings of the 41st International
  Conference on Software Engineering, {ICSE} 2019, {IEEE} / {ACM}, pp 292--303

\bibitem[{Srinivasan(2001)}]{aleph}
Srinivasan A (2001) The {ALEPH} manual. Machine Learning at the Computing
  Laboratory, Oxford University

\bibitem[{Srinivasan et~al(2003)Srinivasan, King, and Bain}]{ashwin:badbk}
Srinivasan A, King RD, Bain M (2003) An empirical study of the use of relevance
  information in inductive logic programming. J Machine Learning Res 4:369--383

\bibitem[{Stahl(1995)}]{stahl:pi}
Stahl I (1995) The appropriateness of predicate invention as bias shift
  operation in {ILP}. Machine Learning 20(1-2):95--117

\bibitem[{Tamaddoni{-}Nezhad et~al(2014)Tamaddoni{-}Nezhad, Bohan, Raybould,
  and Muggleton}]{DBLP:conf/ilp/Tamaddoni-Nezhad14}
Tamaddoni{-}Nezhad A, Bohan D, Raybould A, Muggleton S (2014) Towards machine
  learning of predictive models from ecological data. In: Inductive Logic
  Programming - 24th International Conference, {ILP} 2014, Springer, Lecture
  Notes in Computer Science, vol 9046, pp 154--167

\bibitem[{Vaswani et~al(2017)Vaswani, Shazeer, Parmar, Uszkoreit, Jones, Gomez,
  Kaiser, and Polosukhin}]{vaswani2017attention}
Vaswani A, Shazeer N, Parmar N, Uszkoreit J, Jones L, Gomez AN, Kaiser {\L},
  Polosukhin I (2017) Attention is all you need. In: Advances in neural
  information processing systems, pp 5998--6008

\bibitem[{Wang et~al(2014)Wang, Mazaitis, and Cohen}]{wang2014structure}
Wang WY, Mazaitis K, Cohen WW (2014) Structure learning via parameter learning.
  In: Proceedings of the 23rd {ACM} International Conference on Conference on
  Information and Knowledge Management, {CIKM} 2014, {ACM}, pp 1199--1208

\bibitem[{Wirth(1985)}]{DBLP:books/daglib/0067086}
Wirth N (1985) Algorithms and data structures. Prentice Hall

\bibitem[{Yang et~al(2017)Yang, Yang, and Cohen}]{Cohen_NeuralLP}
Yang F, Yang Z, Cohen WW (2017) Differentiable learning of logical rules for
  knowledge base reasoning. In: {NIPS} 2017

\end{thebibliography}

\end{document}